\documentclass[manuscript,nonacm]{acmart}
\usepackage{booktabs}
\usepackage{graphicx}
\usepackage{multirow,array}
\usepackage{color, xcolor}
\usepackage{soul}
\usepackage[table]{xcolor}

\definecolor{taxD1}{HTML}{F6D7DE}
\definecolor{taxD2}{HTML}{D9ECF5}
\definecolor{taxD3}{HTML}{F7E3BD}
\definecolor{taxD4}{HTML}{E7DFF3}
\definecolor{taxD5}{HTML}{DDECDC}

\newcommand{\dimtag}[2]{%
    \rotatebox[origin=c]{90}{%
        \colorbox{#1}{%
            \scriptsize\bfseries
            \strut\hspace{2pt}#2\hspace{2pt}%
        }%
    }%
}

\newcommand{\currentguidelinecolor}{taxD2}

\newcommand{\guidelinegroup}[2]{%
  \def\currentguidelinecolor{#1}%
  \par\medskip
  \noindent
  \begingroup
    \setlength{\fboxsep}{4pt}%
    \colorbox{#1}{%
      \parbox{\dimexpr\linewidth-2\fboxsep\relax}{%
        \sffamily\bfseries #2%
      }%
    }%
  \endgroup
  \par\nobreak\smallskip
}

\newcommand{\guidelinequestion}[2]{%
  \par\smallskip
  \noindent
  {\bfseries
    #1.\enspace
    \begingroup
      \setulcolor{\currentguidelinecolor}%
      \setul{0.5ex}{1.25pt}%
      \ul{#2}%
    \endgroup
  }%
  \par\nobreak
}

\AtBeginDocument{%
  }

\setcopyright{acmlicensed}
\copyrightyear{2027}
\acmYear{2027}
\acmDOI{XXXXXXX.XXXXXXX}
\acmConference[CHI '27]{The ACM CHI Conference on Human Factors in Computing Systems}{May 10--14, 2027}{Pittsburgh, PA}
\acmISBN{XXXXXXXXXXXXX}

\authorsaddresses{}
\begin{document}

\title[Can Vision-Language Models Analyze Human-Centered Video?]{Can Vision-Language Models Analyze Human-Centered Video? \\ Mapping Model Capabilities and Human–AI Collaborative Workflows}

\author{Xiyuan Shen}
\email{xyshen@cs.washington.edu}
\orcid{0000-0001-9937-7550}
\affiliation{%
  \institution{University of Washington}
  \streetaddress{185 E Stevens Way NE}
  \city{Seattle}
  \state{Washington}
  \country{USA}
}

\author{Jiuyang Lyu}
\orcid{0009-0009-6888-5291}
\affiliation{%
  \institution{Georgia Institute of Technology}
  \streetaddress{225 North Avenue NW}
  \city{Atlanta, Georgia}
  \country{USA}
  \postcode{30332}
}
\email{jlyu45@gatech.edu}

\author{Seokhyun Hwang}
\email{seokhyun@uw.edu}
\orcid{0000-0001-5244-017X}
\affiliation{%
  \institution{University of Washington}
  \streetaddress{Mary Gates Hall, Ste. 370}
  \city{Seattle}
  \state{Washington}
  \country{USA}
}

\author{Huanfen Yao}
\orcid{}
\affiliation{%
  \institution{Google DeepMind}
  \streetaddress{}
  \city{Mountain View, California}
  \country{USA}
  \postcode{}
}
\email{claireyao@google.com}

\author{Shwetak Patel}
\orcid{0000-0002-6300-4389}
\affiliation{%
  \institution{University of Washington}
  \streetaddress{185 E Stevens Way NE}
  \city{Seattle, Washington}
  \country{USA}
  \postcode{98195}
}
\email{shwetak@cs.washington.edu}

\author{Zhihan Zhang}
\orcid{0000-0001-7394-5409}
\affiliation{%
  \institution{University of Washington}
  \streetaddress{185 E Stevens Way NE}
  \city{Seattle, Washington}
  \country{USA}
  \postcode{98195}
}
\email{zzhihan@cs.washington.edu}

\author{Jacob O. Wobbrock}
\orcid{0000-0003-3675-5491}
\affiliation{%
  \institution{University of Washington}
  \streetaddress{Mary Gates Hall, Ste. 370}
  \city{Seattle, Washington}
  \country{USA}
  \postcode{98195}
}
\email{wobbrock@uw.edu}

\hyphenpenalty=10000
\tolerance=2000
\emergencystretch8em
\raggedbottom

\renewcommand{\shortauthors}{Shen et al.}

\begin{abstract}
Video provides a rich record of human behavior, interaction, and situated contexts, offering important evidence for understanding people and conducting human-centered research.
As vision-language models (VLMs) become increasingly capable of analyzing video, they offer opportunities to automate this traditionally human-intensive process. 
Yet a central question remains: \textit{when can VLMs analyze human-centered video independently, and when does reliable analysis still require human involvement?}
To address this question, we first characterize video analysis practices in human-centered research. We systematically analyze all 1,702 CHI 2026 full papers and identify 125 that annotate videos.
Through iterative coding, we derive a five-dimensional taxonomy spanning analytic purpose, viewpoint, phenomenon, reasoning requirement, and annotation authority.
Grounded in recurring annotation tasks captured by this taxonomy, we construct a benchmark of 15 representative tasks from open datasets to map the capabilities and limitations of a general-purpose VLM.
We examine the division of labor between humans and VLMs by comparing three annotation workflows: VLM alone, human alone, and human verification of VLM outputs.
Across tasks, VLM-alone annotation approaches human accuracy on average ($\mathrm{HNS}=97.0$, where 100 denotes human-alone performance), demonstrating substantial potential to automate human-centered video analysis. Human verification achieves the highest accuracy ($\mathrm{HNS}=121.5$) while reducing human annotation time by 48.9\% and monetary cost by 31.3\%--44.5\% relative to human-alone annotation.
Our findings connect real-world human-centered video analysis tasks and current VLM capabilities, and clarify how human-AI collaboration can make VLM-assisted analysis reliable and efficient.

% Video captures some of the richness of human activity, and annotations make it interpretable as HCI research evidence.
% As vision-language models (VLMs) promise to automate video annotation, we need a systematic account of recurring tasks and when model assistance is appropriate. 
% We review all CHI 2026 full papers and identify 125 that annotate videos. 
% Iterative coding yields a five-dimensional taxonomy spanning analytic purpose, viewpoint, phenomenon, reasoning requirement, and annotation authority. 
% Using this taxonomy, we construct a 15-task benchmark from open datasets to assess how a general-purpose VLM performs across recurring HCI annotation tasks. 
% Our results show that VLM-generated annotations achieved accuracy comparable to human annotations ($\mathrm{HNS}=97.0$, where $\mathrm{HNS}=100$ denotes human parity), while human verification of VLM output achieved the highest accuracy ($\mathrm{HNS}=121.5$), reduced human time by 48.9\%, and monetary cost by 31.3\%--44.5\%. 
% We translate these findings into an evidence-based set of guidelines for selecting and validating human-VLM workflows.
\end{abstract}



\maketitle

\section{Introduction}
Video is a rich source of information about human behavior, interaction, and situated contexts. Across human-computer interaction (HCI), researchers use recordings to study how people move, interact, collaborate, use systems, and make sense of technology in context. However, a recording is not self-interpreting. It becomes analyzable evidence only when researchers transform visible activity into structured representations, a process called \textit{video annotation}. Choices about what to annotate, at what granularity, and whose judgment counts shape which aspects of situated activity become visible and, ultimately, which HCI claims the recording can support. Therefore, video annotation is not merely pre-processing, but a consequential methodological act.

Despite this importance, video annotation is usually treated as a project-specific procedure rather than a subject of methodological study in its own right. Researchers develop codebooks, train annotators, and build workflows that combine human coding with computational tools. Since these workflows are tailored to particular behaviors and systems, knowledge about how HCI researchers annotate video remains fragmented. Consequently, the field lacks a systematic review and organizing taxonomy of its recurring annotation practices.

The rapid rise of general-purpose vision-language models (VLMs)~\cite{alayrac2022flamingo} makes addressing this gap both timely and urgent. 
VLMs can interpret video in response to natural-language instructions and return flexible natural-language or structured outputs, potentially supporting varied annotation tasks without requiring a task-specific model. This flexibility is particularly relevant to HCI, whose research spans human bodies, tangible objects, digital interfaces, and social interactions, but the reliability of VLMs may still vary substantially across these and other research uses.

Determining whether VLMs can meaningfully support HCI video annotation requires looking beyond conventional video-understanding benchmarks. Although such benchmarks measure question answering, captioning, and reasoning~\cite{patraucean2023perception, li2024mvbench, liu2024tempcompass}, they are organized around AI capabilities rather than the methodological and workflow roles of annotation in HCI. They therefore offer limited guidance on when VLM output is suitable as research evidence, or how it should be incorporated into annotation workflows. We address this gap by characterizing current HCI practices, identifying recurring tasks in which human judgment remains central, and evaluating a general-purpose VLM across tasks and workflows. This approach leads to two research questions:

\noindent\textbf{RQ1. How is video annotation commonly used in recent HCI research?}
To answer RQ1, we analyze all 1,702 full papers published at CHI 2026 and identify 125 that annotate video for research purposes. Through iterative coding, we develop a taxonomy with five methodological dimensions: (1) \textit{analytic purpose}, or the role annotation serves in the research workflow; (2) \textit{video viewpoint}, or the perspective from which visual evidence is available; (3) \textit{annotated phenomenon}, or the aspect of the visual record treated as analytically relevant; (4) \textit{reasoning requirement}, or the inference needed to produce the annotation; and (5) \textit{annotation authority}, or the source whose labels are treated as authoritative.

\noindent\textbf{RQ2. Across recurring video annotation tasks in HCI, when do general-purpose VLMs work well, and how should they be used in annotation workflows?}
To answer RQ2, we retain 74 papers that include human-produced, human-machine hybrid, or general-purpose VLM annotation streams. We represent each paper as a profile across our taxonomy dimensions and cluster these profiles, yielding five recurring task families: \textit{Long-Horizon Interaction Interpretation}, \textit{Egocentric Spatial Grounding}, \textit{Interface Workflow Understanding}, \textit{Fine-Grained Embodied Behavior Recognition}, and \textit{Communicative Signal Interpretation}. We instantiate these families as 15 representative tasks drawn from open datasets, comprising 1,459 task instances. We then compare three annotation workflows: a general-purpose VLM operating alone, humans annotating from scratch, and humans verifying VLM-generated labels. We measure accuracy using Human-Normalized Score (HNS)~\cite{volodymyr2015human,srivastava2022beyond}, for which 0 represents chance performance and 100 represents human-alone performance on a task. We also measure human annotation time and monetary cost.

In our benchmark, VLM-alone annotation performed comparably to human-alone annotation on average (mean $\mathrm{HNS}=97.0$), while human verification of VLM output achieved the highest average performance (mean $\mathrm{HNS}=121.5$).
Across the tasks we evaluated, performance depended less on video viewpoint or temporal horizon than on the evidence and judgment required by the label. The VLM performed well when clear operational criteria could be applied to salient actions, objects, or interfaces. It struggled when annotation required inferring subtle states or applying specialized conventions. In particular, its tendency to miss failures and other low-salience events may lead researchers to underestimate breakdowns.
VLM assistance also reduced human annotation time and monetary cost. Verification required 48.9\% less human annotation time than human-alone annotation, while the two VLM-based workflows reduced estimated costs by 31.3\%--95.6\%.
We distill these findings into guiding questions for operationalizing annotation tasks, validating VLM performance, and selecting an appropriate human-VLM annotation workflow.

We make the following two contributions:
\begin{itemize}
\item \textbf{A taxonomy of video annotations in HCI.} Through a systematic literature review, we develop a five-dimensional taxonomy that organizes video annotation practices across HCI research.
\item \textbf{A taxonomy-grounded evaluation of VLM annotation workflows.} We compare VLM-alone annotation, human-alone annotation, and human verification of VLM output. We characterize the VLM's strengths and limitations, deriving guiding questions for operationalizing annotation tasks and selecting human-VLM workflows.
\end{itemize}

\section{Related Work}
We situate our work in three areas: reviews and taxonomies of VLMs in applied systems, benchmarks for video understanding, and research on foundation models as research infrastructure in HCI.

\subsection{Literature Reviews and Taxonomies of VLMs in Applied Systems}
Existing reviews of vision-language models (VLMs) and related vision-language systems follow two main orientations: model-centered and application-centered. Model-centered surveys organize the field by architecture, pretraining data, adaptation strategies, and evaluation benchmarks, while treating downstream uses primarily as capability categories~\cite{li2024multimodal, yao2025survey}. They trace a shift from specialized models toward general-purpose multimodal assistants that support open-ended visual dialogue~\cite{liu2023visual}, multimodal content generation~\cite{wu2023next}, embodied reasoning and action~\cite{driess2023palm}, and biomedical interpretation~\cite{tu2024towards}. This orientation explains how general-purpose capabilities are constructed, but pays less attention to how VLM outputs are situated within particular systems and use cases.
Application-centered reviews instead organize research by domain-specific tasks and system functions. Video surveys distinguish captioning and description~\cite{aafaq2019video}, question answering~\cite{zhong2022video}, LLM-based retrieval, and temporal grounding~\cite{tang2025video}. Reviews of vision-language navigation classify tasks by environment, instruction structure, and interaction protocol~\cite{park2023visual}, while autonomous-driving taxonomies organize VLM uses across perception, navigation, and data generation~\cite{zhou2024vision}. These frameworks clarify how VLM capabilities serve functional roles within specific domains.

Related human-computer interaction (HCI) reviews, instead, move the analysis from the model to the interactive system. Hu et al.'s survey of vision-based multimodal interfaces~\cite{hu2025vision}, including VLM-enabled systems, organizes prior work by sensed context, input modality, processing and integration choices, evaluation, and application domain. This emphasis on situated use also appears in HCI studies that use VLM outputs to reconstruct workflows from screen recordings~\cite{10.1145/3772318.3790294}, characterize videos in recommendation feeds~\cite{10.1145/3772318.3790311}, and propose contextual labels for social-touch events that human coders subsequently verify~\cite{10.1145/3772318.3791605}. In each case, the VLM's role depends on how its outputs enter a larger workflow.
Taken together, prior taxonomies organize VLMs by model design and application domain, but do not examine video annotation as a research methodology through which HCI studies turn recordings into research evidence. We address this gap through a literature review and the development of a practice-grounded taxonomy.

\subsection{Benchmarking Vision-Language Models for Video Understanding}
In artificial intelligence (AI), video benchmarks test multimodal capabilities that image-based evaluation cannot capture. The representative Perception Test~\cite{patraucean2023perception} combines question answering with spatiotemporal localization to assess multimodal perception and reasoning. MVBench covers 20 tasks spanning video perception and cognition~\cite{li2024mvbench}. TempCompass targets motion, state change, and event order across multiple task formats~\cite{liu2024tempcompass}, and Video-MME broadens evaluation across domains, lengths, and modalities~\cite{fu2025video}. Other benchmarks examine long-context reasoning and output reliability~\cite{mangalam2023egoschema, wu2024longvideobench,li2025vidhalluc}. Together, these benchmarks show that video understanding comprises distinct abilities requiring separate evaluation.

These benchmarks are indispensable for model diagnosis and comparison, but primarily operationalize video understanding as model capability. HCI video annotation serves a different purpose: researchers apply study-specific codebooks to transform recorded activity into evidence for particular research questions, often using labels with subjective or ambiguous boundaries. Human annotation is therefore not merely a ceiling for model performance, but an alternative workflow that must be compared in accuracy, time, and monetary cost. Following calls to evaluate models within downstream contexts and human requirements~\cite{liao2023rethinking}, our paper compares VLM-alone annotation, human-alone annotation, and human verification of VLM output. We therefore evaluate when VLMs can support video annotation and how workflow choice affects annotation accuracy and efficiency.

\subsection{Foundation Models as Research Infrastructure in HCI}
Foundation models are beginning to function as research infrastructure in HCI: beyond powering interactive systems and serving as objects of study, they increasingly participate in producing research evidence. A systematic review of 153 CHI papers identifies research tooling as a distinct role for LLMs, while a survey of 816 researchers documents their use throughout the research process~\cite{pang2025understanding,liao2024llms}. Once model outputs support empirical claims, the procedures used to produce and review them become part of the research method.
These research roles vary with the data being analyzed. For text, LLooM helps researchers identify recurring concepts and apply them across documents~\cite{lam2024concept, schroeder2025large}. VLMs extend this role to visual data, supporting both automated annotation and hybrid workflows in which humans review model-proposed labels. HCI studies have applied these approaches to codebook-based video analysis, behavioral measurement, and the characterization of video collections~\cite{whitehead2025utilizing,10.1145/3772318.3790294,10.1145/3772318.3790311,10.1145/3772318.3791605}. Related machine-learning research similarly uses VLM-generated descriptions and judgments to construct datasets and evaluate other models at scale~\cite{chen2024sharegpt4video, yeh2024t2vs, ku2024viescore}. In each case, model outputs become inputs to subsequent analysis and thus shape what is reported as evidence.

Human review is therefore important, but does not by itself ensure valid annotations. AI suggestions may direct reviewers toward likely errors, but can also narrow their interpretations or be accepted without sufficient scrutiny~\cite{wang2024human, gao2023coaicoder, buccinca2021trust}. Agreement and efficiency thus provide only partial evidence of workflow quality. Evaluation must determine whether reviewers can recognize and correct model errors, while also accounting for the time and cost of verification. Existing studies typically validate model-assisted annotation on a single task or dataset, offering limited evidence about variation across HCI video-annotation practices. Reproducibility is also challenging because proprietary models and interfaces change over time~\cite{kosch2024risk}. We address these limitations by systematically evaluating VLMs as annotation infrastructure and comparing workflows across various tasks.

\section{Characterizing Video Annotation in HCI}\label{sec:taxonomy}
To characterize how video annotation is typically used in recent human–computer interaction (HCI) research (RQ1), we conducted a systematic review of the CHI 2026 proceedings and developed a taxonomy through iterative coding. The taxonomy captures recurring methodological choices in how video is transformed into research evidence.

\subsection{Data}
We assembled a corpus of all 1,702 full papers published in the CHI 2026 proceedings. We selected CHI for two reasons. 
First, CHI is the flagship international conference in HCI. Its papers undergo rigorous peer review, and prior meta-research has used its proceedings as a broad lens for characterizing intellectual and methodological trends in the field~\cite{liu2014chi,linxen2021weird}. Second, CHI spans the methodological and topical breadth of HCI. Its proceedings therefore cover diverse settings in which video annotation arises. Focusing on the most recent proceedings captures current practices, including emerging uses of VLM-assisted annotation. Our sample is intended to generate a taxonomy rather than exhaustively represent all HCI research. More specialized venues may contain additional annotation practices.

We followed an adapted PRISMA procedure~\cite{page2021prisma}. We restricted both keyword filtering and manual screening to each paper’s title, keywords, abstract, and introduction. We first retained papers in which these sections contained “video” together with at least one annotation-related term, including variants of “annotation,” “labeling/labelling,” and “coding.” One author then manually reviewed the same sections and retained papers that described annotating video content for a substantive research purpose. This process yielded a final corpus of 125 papers.

\begin{figure}[]
    \centering
    \includegraphics[width=0.7\linewidth]{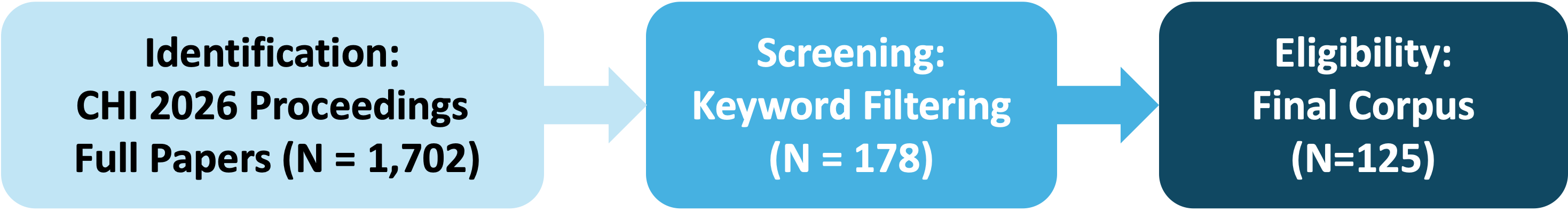}
    \caption{PRISMA-style flow diagram showing the initial corpus (N = 1,702), papers retained after keyword filtering of titles, keywords, abstracts, and introductions (N = 178), and the final corpus after manual screening (N = 125).}
    \Description{A left-to-right flow diagram of paper selection. The initial corpus contains 1,702 CHI 2026 full papers, keyword filtering retains 178 papers, and manual screening produces the final corpus of 125 papers.}
    \label{fig:PRISMA}
\end{figure}

\subsection{Analysis}
To analyze the 125 papers, three authors developed the codebook through four iterative rounds. In the first round, they jointly reviewed 20 papers and discussed recurring variations in video annotation practice to establish an initial set of dimensions, categories, definitions, and coding rules. In each of the subsequent three rounds, the coders independently coded a randomly selected set of 10 papers. They then compared their codes, refined category definitions, merged or split categories, specified rules for boundary cases, and resolved disagreements through consensus. Before reconciliation, we calculated Krippendorff's alpha~\cite{krippendorff1970estimating} on the independently assigned codes and used the results to identify ambiguities requiring further refinement.

After the fourth round, the codebook stabilized around five methodological dimensions: analytic purpose, video viewpoint, annotated phenomenon, reasoning requirement, and annotation authority. 
For each paper, we coded the annotation stream being used substantively in the research rather than every tool or preprocessing step in the workflow. Each stream received one category in each dimension. We coded up to two streams only when a paper produced and used distinct annotation streams for separate research purposes, such as across separate studies. The final reliability values were $\alpha_{purpose}=1.00$, $\alpha_{viewpoint}=.896$, $\alpha_{phenomenon}=.817$, $\alpha_{reasoning}=.821$, and $\alpha_{authority}=.964$. All exceeded Krippendorff's recommended threshold of 0.80 for drawing reliable conclusions from coded data~\cite{lombard2002content}.

We did not use LLMs during coding. All codes reflect the interpretive judgments of the three coding authors, whose academic backgrounds span input technologies, health, virtual reality, human-AI interaction, agentic systems, and sustainability. This breadth supports construct validity by grounding the taxonomy in methodological expertise rather than surface textual features. Meanwhile, our perspectives shaped the distinctions we considered salient, and researchers from other scholarly traditions might partition the space differently.

\section{A Taxonomy of HCI Video Annotation Practices}
To answer RQ1, we organize recurring HCI video annotation practices along five dimensions (Table~\ref{tab:taxonomy}). These dimensions characterize why annotation is performed, the viewpoint from which evidence is available, what is labeled, what reasoning the label requires, and whose judgment determines the final annotation. 
Together, they represent each annotation stream as a five-part methodological profile and describe how studies transform video into research evidence. 
The complete codebook and coding results are provided in the supplementary materials.

\subsection{Analytic Purpose (D1)}
Analytic purpose classifies the role that video annotation serves within a paper's research workflow. The classification is determined by how the resulting annotations are used.

\textbf{R1. Empirical outcome analysis} (61 papers, 48.8\%): 
Annotation supports empirical claims about users, systems, and their interactions by describing, measuring, or interpreting observable behavior. Video makes situated behavior available for systematic analysis. Accessibility studies use coded video to reveal interaction strategies and barriers that inform accessible interaction design~\cite{10.1145/3772318.3791184, 10.1145/3772318.3790437, 10.1145/3772318.3791293, 10.1145/3772318.3791708,10.1145/3772318.3791927}. Studies of embodied interaction in extended reality (XR) use video to compare movement, gesture, and spatial coordination~\cite{10.1145/3772318.3790651, 10.1145/3772318.3790491, 10.1145/3772318.3790512}. Human-robot and animal-computer interaction studies analyze behavioral responses to explain coordination with agents and devices~\cite{10.1145/3772318.3791828, 10.1145/3772318.3793419, 10.1145/3772318.3790842}.

\textbf{R2. Cross-modal reference labeling} (22 papers, 17.6\%): 
Annotation establishes ground-truth or reference labels used to train, validate, or evaluate what another sensing modality or computational system is expected to infer. This role turns video-based observations into a common standard for assessing indirect measurements and model outputs. First, a large number of papers use video capture as the reference stream for another sensing channel. These studies most often label hand and body motion to assess estimates derived from IMUs, acoustic sensing, pressure, or wearable cameras~\cite{10.1145/3772318.3790932, 10.1145/3772318.3790493, 10.1145/3772318.3790626, 10.1145/3772318.3791273, 10.1145/3772318.3790607}. Second, researchers use human annotations to benchmark screen and video-understanding systems. These labels support evaluations of GUI interaction prediction~\cite{10.1145/3772318.3790283, 10.1145/3772318.3790294, 10.1145/3772318.3791178}, temporal event segmentation, and procedural step extraction~\cite{10.1145/3772318.3790679, 10.1145/3772318.3790494, 10.1145/3772318.3790774}.

\textbf{R3. Content and resource construction} (6 papers, 4.8\%): Annotation transforms video into structured semantic content, annotated corpora, or design materials for system design and development. Four papers use formative corpus analysis to derive taxonomies or design references that guide system development~\cite{10.1145/3772318.3791238, 10.1145/3772318.3790715, 10.1145/3772318.3790830, 10.1145/3772318.3791652}. The other two papers construct labeled datasets for reuse and model training~\cite{10.1145/3772318.3791220, 10.1145/3772318.3790565}.

\textbf{R4. System-embedded interpretation} (56 papers, 44.8\%): Annotation operates as a functional component of a running interactive or AI system. It converts video into an operational representation of the user, task, or interaction state to provide feedback or generate content. Here, machine interpretation is integrated directly into system operation. First, live systems interpret incoming frames under latency constraints. For example, scene grounding turns camera streams into spoken descriptions and navigation guidance~\cite{10.1145/3772318.3791308, 10.1145/3772318.3791322, 10.1145/3772318.3790589, 10.1145/3772318.3790449}. Hand pose and object annotations drive interactions in XR and wearable interfaces~\cite{10.1145/3772318.3790938, 10.1145/3772318.3791028, 10.1145/3772318.3790626, 10.1145/3772318.3790607}. User-state interpretation converts inferred affect into adaptive interventions~\cite{10.1145/3772318.3791399, 10.1145/3772318.3790990, 10.1145/3772318.3791488, 10.1145/3772318.3791156} and personalized coaching~\cite{10.1145/3772318.3791652, 10.1145/3772318.3790634, 10.1145/3772318.3790774}. Second, offline pipelines transform recorded video into structured content for interaction, including AR tutorials, workflow recommendations, and generated media~\cite{10.1145/3772318.3790715, 10.1145/3772318.3791572, 10.1145/3772318.3791162, 10.1145/3772318.3790269, 10.1145/3772318.3791532}.

%%%%%%%%%%%%%%%%%%%%%%%%%%TABLE
\begin{table}[!t]
\centering
\caption{Taxonomy of video annotation practices in HCI.}
\Description{Taxonomy of HCI video annotation practices. The table defines five dimensions and their codes: four analytic purposes, five video viewpoints, seven types of annotated phenomena, four reasoning requirements, and four sources of annotation authority. Each code is paired with a concise operational definition.}
\label{tab:taxonomy}
\footnotesize
\setlength{\tabcolsep}{4pt}
\renewcommand{\arraystretch}{1.10}

\begin{tabular}{
    @{}
    >{\centering\arraybackslash}p{0.046\linewidth}
    >{\raggedright\arraybackslash}p{0.37\linewidth}
    >{\raggedright\arraybackslash}p{0.54\linewidth}
    @{}
}
\toprule
& \textbf{Code} & \textbf{Definition} \\
\midrule

% D1: Analytic purpose
& \multicolumn{2}{@{}l@{}}{%
    \textit{\textbf{D1. Analytic purpose:}
    What role does annotation serve in the research?}%
} \\
\cmidrule(lr){2-3}

\multirow[c]{4}{*}{\dimtag{taxD1}{Purpose}}
& \textbf{R1. Empirical outcome analysis}
& Supports empirical claims by measuring or interpreting observed behavior. \\

& \textbf{R2. Cross-modal reference labeling}
& Provides reference labels for training or evaluating another sensor or computational model. \\

& \textbf{R3. Content and resource construction}
& Converts video into reusable datasets, taxonomies, or design resources. \\

& \textbf{R4. System-embedded interpretation}
& Supplies video-derived representations used by an interactive or AI system. \\

\addlinespace[5pt]

% D2: Video viewpoint
& \multicolumn{2}{@{}l@{}}{%
    \textit{\textbf{D2. Video viewpoint:}
    From what camera perspective is visual evidence available?}%
} \\
\cmidrule(lr){2-3}

\multirow[c]{5}{*}{\dimtag{taxD2}{Viewpoint}}
& \textbf{V1. External scene view}
& Records the observed person, object, or setting from an external camera. \\

& \textbf{V2. Egocentric or participant-worn}
& Records activity from a camera worn by the participant. \\

& \textbf{V3. Object-mounted}
& Records the changing view of a moving robot, vehicle, animal, or instrument. \\

& \textbf{V4. Screen or 2D digital interaction recording}
& Records activity within a 2D interface, such as a software workflow or gameplay. \\

& \textbf{V5. Synthetic or computationally generated}
& Presents rendered, simulated, generated, or computationally reconstructed content. \\

\addlinespace[5pt]

% D3: Annotated phenomenon
& \multicolumn{2}{@{}l@{}}{%
    \textit{\textbf{D3. Annotated phenomenon:}
    What in the video is being labeled?}%
} \\
\cmidrule(lr){2-3}

\multirow[c]{7}{*}{\dimtag{taxD3}{Phenomenon}}
& \textbf{P1. Human movement, posture, and touch}
& Labels body configuration, movement, trajectory, or physical contact. \\

& \textbf{P2. Attention, affect, and mental-state cues}
& Uses gaze, facial, or bodily cues to characterize attention, affect, or related states. \\

& \textbf{P3. Interface and digital content}
& Labels visible interface or media elements, states, actions, and transitions. \\

& \textbf{P4. Scene objects and environmental context}
& Labels objects, text, spatial relations, infrastructure, or environmental conditions. \\

& \textbf{P5. Visually grounded human-human communication}
& Relates visible cues to communicative meaning or interpersonal coordination. \\

& \textbf{P6. Passive object and device interaction}
& Labels contact with and manipulation of non-agentic physical artifacts. \\

& \textbf{P7. Active agent, robot, and pet interaction}
& Labels exchanges with entities whose actions or responses contribute to the interaction. \\

\addlinespace[5pt]

% D4: Reasoning requirement
& \multicolumn{2}{@{}l@{}}{%
    \textit{\textbf{D4. Reasoning requirement:}
    What visual reasoning is needed to produce the label?}%
} \\
\cmidrule(lr){2-3}

\multirow[c]{4}{*}{\dimtag{taxD4}{Reasoning}}
& \textbf{M1. Scene-level or target-given judgment}
& Judges a supplied scene, clip, or target without locating it in space or time. \\

& \textbf{M2. Spatial-reference grounding}
& Binds a label to a region, object, landmark, or coordinate within a frame. \\

& \textbf{M3. Local temporal reasoning}
& Detects, segments, or tracks change within a bounded interval. \\

& \textbf{M4. Global sequence reasoning}
& Integrates separated events to infer order, dependencies, progress, or interaction history. \\

\addlinespace[5pt]

% D5: Annotation authority
& \multicolumn{2}{@{}l@{}}{%
    \textit{\textbf{D5. Annotation authority:}
    Whose judgment determines the final annotation?}%
} \\
\cmidrule(lr){2-3}

\multirow[c]{4}{*}{\dimtag{taxD5}{Authority}}
& \textbf{A1. Researcher or domain-expert annotation}
& Treats labels produced by researchers or domain experts as authoritative. \\

& \textbf{A2. Participant or stakeholder annotation}
& Treats first-person, situated, or aggregated stakeholder judgments as authoritative. \\

& \textbf{A3. Automatic computational annotation}
& Uses rule- or model-generated labels without substantive human correction. \\

& \textbf{A4. Human-machine hybrid annotation}
& Combines machine inference with substantive human input, review, or correction. \\

\bottomrule
\end{tabular}
\end{table}
%%%%%%%%%%%%%%%%%TABLE ENDS

\subsection{Video Viewpoint (D2)}
Video viewpoint classifies the camera perspective and source context through which the annotated phenomenon becomes visible.

\textbf{V1. External scene view} (70 papers, 56.0\%): Video is recorded from a camera positioned outside the person or object. This category includes fixed, handheld, and multi-camera recordings. Most papers use researcher-controlled cameras to establish a consistent observer perspective on embodied activity. Cameras placed around laboratory and field settings make human activities and spatial relations visible for systematic comparison across participants and conditions~\cite{10.1145/3772318.3790565, 10.1145/3772318.3793422, 10.1145/3772318.3791969, 10.1145/3772318.3790741, 10.1145/3772318.3790618}. This configuration also supports situated observation of robots in transit spaces, drivers in vehicles, and children in food gardens~\cite{10.1145/3772318.3791828, 10.1145/3772318.3790972, 10.1145/3772318.3791719}. Participant-facing webcams provide a narrower external view of seated participants’ faces and upper bodies for annotating affect, attention, and facial expression~\cite{10.1145/3772318.3791488, 10.1145/3772318.3791208, 10.1145/3772318.3790990, 10.1145/3772318.3791927, 10.1145/3772318.3791941}.

\textbf{V2. Egocentric or participant-worn} (47 papers, 37.6\%): Video is captured by a participant-worn camera, providing a perceptual or body-centered view of activity. XR headsets such as Meta Quest, HoloLens, and Apple Vision Pro support annotations of hands, objects, and task states that enable manipulation, AR guidance, and interaction with embodied agents \cite{10.1145/3772318.3790384, 10.1145/3772318.3791059, 10.1145/3772318.3790494, 10.1145/3772318.3790288, 10.1145/3772318.3790938}. Non-headset cameras mounted on the wrist, ears, or head \cite{10.1145/3772318.3790626, 10.1145/3772318.3791322, 10.1145/3772318.3790607, 10.1145/3772318.3790774} are especially common in accessibility research, where annotations transform surrounding scenes and hand-object activity into descriptions or guidance~\cite{10.1145/3772318.3791308, 10.1145/3772318.3790589, 10.1145/3772318.3791659, 10.1145/3772318.3790944, 10.1145/3772318.3790817}. Additionally, eye-tracking glasses combine scene video with gaze, enabling researchers to study visual attention~\cite{10.1145/3772318.3793419, 10.1145/3772318.3791184, 10.1145/3772318.3793422}.

\textbf{V3. Object-mounted} (9 papers, 7.2\%): Video is captured by a camera attached to a moving non-participant entity, such as a robot, vehicle, or instrument. This alignment makes the platform’s changing field of view available for annotating the environment. Robot- and vehicle-mounted cameras turn local scenes into structured evidence for sensing, navigation, and user-facing assistance~\cite{10.1145/3772318.3790679, 10.1145/3772318.3790345, 10.1145/3772318.3791828, 10.1145/3772318.3791195, 10.1145/3772318.3791220}. Other mounts provide task-specific views, including drone and canine-camera footage for search-and-rescue sensemaking, and endoscopic video for instrument-mediated interaction~\cite{10.1145/3772318.3791523, 10.1145/3772318.3790904}.

\textbf{V4. Screen or 2D digital interaction recording} (14 papers, 11.2\%): Video records activity on a 2D digital interface, including screen recordings, software workflow videos, and gameplay capture. Annotation translates visible interface states and transitions into representations of user behavior or system state. First, researchers code screen recordings to reconstruct how people work with software~\cite{10.1145/3772318.3790294, 10.1145/3772318.3791776, 10.1145/3772318.3790318, 10.1145/3772318.3791448, 10.1145/3772318.3791293}. Second, annotations of interface elements, actions, and state transitions provide training or evaluation data for GUI understanding and game-content detection~\cite{10.1145/3772318.3791958, 10.1145/3772318.3790283, 10.1145/3772318.3791178}.

\textbf{V5. Synthetic or computationally generated} (5 papers, 4.0\%): Video consists of AI-generated or computationally reconstructed visual content. Three papers analyze generated content directly, including events and object motion in animated graphics, virtual-camera practices in VRChat streams, and Unity replays reconstructed from logged VR sessions~\cite{10.1145/3772318.3791162, 10.1145/3772318.3790858, 10.1145/3772318.3790591}. Two others use generated video as controlled material alongside captured footage~\cite{10.1145/3772318.3790501, 10.1145/3772318.3791572}.

\subsection{Annotated Phenomenon (D3)}
This dimension identifies which phenomenon in the visual record a study selects as analytically relevant. 

\textbf{P1. Human movement, posture, and touch} (34 papers, 27.2\%): Annotation captures visible body configurations, including movement, posture, trajectory, and physical contact. It turns bodily behavior into spatial and temporal representations. Hand and finger configuration is the most common. Studies annotate joint landmarks, hand meshes, fingertip positions, and grasp postures to train or evaluate input-sensing and pose-estimation methods~\cite{10.1145/3772318.3790493, 10.1145/3772318.3790932, 10.1145/3772318.3790626, 10.1145/3772318.3791028, 10.1145/3772318.3790836}. Whole-body annotations record body trajectories and head and limb positions to support comparisons of postures across XR and physical settings~\cite{10.1145/3772318.3790992, 10.1145/3772318.3790486, 10.1145/3772318.3790805, 10.1145/3772318.3790512, 10.1145/3772318.3790614}. Motor-skill and rehabilitation studies code postures and joint angles against clinical references to support performance assessment and corrective feedback~\cite{10.1145/3772318.3790634, 10.1145/3772318.3791652, 10.1145/3772318.3790808, 10.1145/3772318.3790414, 10.1145/3772318.3791156}. Gesture and facial-movement annotations represent signing mechanics, gesture forms, and facial kinematics for characterizing movement patterns~\cite{10.1145/3772318.3790774, 10.1145/3772318.3791927, 10.1145/3772318.3790345, 10.1145/3772318.3791941, 10.1145/3772318.3791122}.

\textbf{P2. Attention, affect, and mental-state cues} (29 papers, 23.2\%): Annotation uses facial, gaze, and bodily cues to characterize attention, affect, and related mental states. This process makes latent states empirically tractable and supplies state estimates for adaptive systems. One pattern operationalizes how visual attention is allocated across people, objects, and interface regions through eye fixations, scanpaths, and glance distributions~\cite{10.1145/3772318.3791178, 10.1145/3772318.3790434, 10.1145/3772318.3790738, 10.1145/3772318.3791192, 10.1145/3772318.3791758}. A second pattern infers affective states from facial expressions. These annotations range from discrete emotion categories to composite states such as fatigue, cognitive load and engagement~\cite{10.1145/3772318.3791488, 10.1145/3772318.3791208, 10.1145/3772318.3790745, 10.1145/3772318.3791153, 10.1145/3772318.3790972, 10.1145/3772318.3791481}.

\textbf{P3. Interface and digital content} (13 papers, 10.4\%): Annotation captures the visible content and evolution of digital interfaces. By structuring what appears and changes on screen, it supports analyses of software use and enables computational systems. A prominent strand reconstructs graphical user interface (GUI) workflows by labeling interface elements, ordered user actions, and usability problems in screen recordings~\cite{10.1145/3772318.3790294, 10.1145/3772318.3790283, 10.1145/3772318.3791293, 10.1145/3772318.3790536, 10.1145/3772318.3790318}. A complementary strand characterizes rendered media, including social-media content, deepfake artifacts, game entities, animated motion graphics, and virtual-camera framing~\cite{10.1145/3772318.3790501, 10.1145/3772318.3791958, 10.1145/3772318.3790311, 10.1145/3772318.3791162, 10.1145/3772318.3790858}.

\textbf{P4. Scene objects and environmental context} (27 papers, 21.6\%): Annotation structures a visual scene by identifying objects, spatial relations, infrastructure, and environmental conditions. A major strand supports accessibility and mobility. Studies identify objects and printed text, reconstruct spatial layouts, and characterize roads and outdoor infrastructure. These annotations support blind and low-vision access as well as transportation and public-safety analysis~\cite{10.1145/3772318.3790449, 10.1145/3772318.3791308, 10.1145/3772318.3791322, 10.1145/3772318.3790589, 10.1145/3772318.3791309, 10.1145/3772318.3791099}. Another strand converts scene structure into anchors for interactive content. XR and media-authoring pipelines identify objects and estimate their geometry or pose so that instructions and generated content can be aligned with the physical environment~\cite{10.1145/3772318.3790938, 10.1145/3772318.3790715, 10.1145/3772318.3791929, 10.1145/3772318.3791202, 10.1145/3772318.3791532}.

\textbf{P5. Visually grounded human–human communication} (17 papers, 13.6\%): Annotation captures interpersonal communication by relating visible cues such as gesture, gaze, and posture to speech. These annotations reveal what particular signals convey and how participants coordinate an exchange. For example, co-speech studies align gestures with utterances to determine how the two modalities distribute meaning~\cite{10.1145/3772318.3790618, 10.1145/3772318.3790641, 10.1145/3772318.3790491, 10.1145/3772318.3790741}. Accessibility studies connect observable signals to intended messages and interactional functions~\cite{10.1145/3772318.3791273, 10.1145/3772318.3790944, 10.1145/3772318.3791708, 10.1145/3772318.3791520}. Another strand tracks how communication develops across turns. Studies annotate joint attention, response timing, turn-taking, and coordinated task actions to examine how participants establish and maintain shared activity~\cite{10.1145/3772318.3791267, 10.1145/3772318.3790922, 10.1145/3772318.3791238, 10.1145/3772318.3790284, 10.1145/3772318.3791399}.

\textbf{P6. Passive object and device interaction} (16 papers, 12.8\%): Annotation captures how people or animals handle passive physical artifacts through contact and manipulation. The labels record where contact occurs, how an artifact is held, and which actions are performed over time. One group of studies focuses on grasp and tactile contact. Researchers label these events and exploration strategies to characterize device use or evaluate sensing methods~\cite{10.1145/3772318.3790565, 10.1145/3772318.3790384, 10.1145/3772318.3790607, 10.1145/3772318.3790904, 10.1145/3772318.3790437}. A complementary group represents artifact use as sequences of goal-directed actions. Labels for steps such as removing, aligning, inserting, and operating components support tutorial generation and physical assistance~\cite{10.1145/3772318.3790715, 10.1145/3772318.3790494, 10.1145/3772318.3790817, 10.1145/3772318.3791059}. This approach also captures extended engagement with animal enrichment devices~\cite{10.1145/3772318.3790842, 10.1145/3772318.3791644}.

\textbf{P7. Active agent, robot, and pet interaction} (11 papers, 8.8\%): Annotation characterizes interaction with robots, virtual agents, avatars, animals, and other interactive artifacts. It traces how actions and responses develop over time, including moments of coordination, breakdown, and adaptation. Studies commonly examine responses to robots and virtual agents, including path negotiation, reactions to malfunctions, turn-taking, facilitation, and interpersonal distance~\cite{10.1145/3772318.3791828, 10.1145/3772318.3793419, 10.1145/3772318.3790947, 10.1145/3772318.3791910, 10.1145/3772318.3791241}. Other studies follow social and affective exchanges involving animals, interactive artifacts, and generative systems~\cite{10.1145/3772318.3790830, 10.1145/3772318.3790842, 10.1145/3772318.3791969, 10.1145/3772318.3791776, 10.1145/3772318.3791708}. 

\subsection{Reasoning Requirement (D4)}
This dimension characterizes the visual reasoning needed to produce an annotation. The categories distinguish semantic judgment of a supplied scene or target (M1), spatial grounding at one time point (M2), temporal interpretation within a bounded interval (M3), and sequence-level integration across a longer recording (M4).

\textbf{M1. Scene-level or target-given judgment} (13 papers, 10.4\%): Annotation assigns a visual property, category, or description to an overall scene or a target specified in advance. Because the annotation unit and target are already given, the annotator makes a semantic or evaluative judgment about what is shown. Two practices recur. One assigns a categorical or ordinal label to a complete item or pre-segmented clip \cite{10.1145/3772318.3791039, 10.1145/3772318.3791481, 10.1145/3772318.3791488, 10.1145/3772318.3791308}. The other applies a rubric to visual material, including judgments of image quality, product-caption quality, street-scene attributes, and holistic perceptual responses \cite{10.1145/3772318.3791292, 10.1145/3772318.3791309, 10.1145/3772318.3790918, 10.1145/3772318.3790836}.

\textbf{M2. Spatial-reference grounding} (34 papers, 27.2\%): Annotation links visual information to a location within a single frame. It answers \textit{where} or \textit{which one} by binding a label to a region, object, or coordinate. This spatial grounding makes visible targets addressable for different downstream applications. First, body- and hand-tracking studies extract per-frame landmarks, meshes, and contact points to represent pose and physical contact \cite{10.1145/3772318.3790932, 10.1145/3772318.3790626, 10.1145/3772318.3790493, 10.1145/3772318.3791156, 10.1145/3772318.3790414}. In another application, assistive, wearable, and vehicle-mounted systems localize objects and hazards to support access and safety \cite{10.1145/3772318.3791195, 10.1145/3772318.3790288, 10.1145/3772318.3791322, 10.1145/3772318.3790589, 10.1145/3772318.3791523, 10.1145/3772318.3791929}. Attention studies map gaze to areas of interest, reconstructed mesh faces, or screen coordinates to identify what a person is viewing \cite{10.1145/3772318.3790738, 10.1145/3772318.3791192, 10.1145/3772318.3791339, 10.1145/3772318.3790629}. Finally, interactive systems ground linguistic references in visible objects, allowing speech or gesture to specify the intended referent~\cite{10.1145/3772318.3790938, 10.1145/3772318.3790604, 10.1145/3772318.3790817}.

\textbf{M3. Local temporal reasoning} (59 papers, 47.2\%): Annotation detects visual change within a bounded interval. It draws on adjacent frames or short windows to determine when an event occurs, how long it lasts, or how a state changes. The category spans four common workflows. First, researchers apply codebooks to mark discrete behaviors and when they occur in animal-computer interaction, driving research, human-robot interaction, and social communication~\cite{10.1145/3772318.3790842, 10.1145/3772318.3791644, 10.1145/3772318.3790972, 10.1145/3772318.3793419, 10.1145/3772318.3791605}. Second, temporal segmentation marks the onset and offset of communicative movements and interactive gestures~\cite{10.1145/3772318.3791273, 10.1145/3772318.3790491, 10.1145/3772318.3790774, 10.1145/3772318.3790618, 10.1145/3772318.3791927, 10.1145/3772318.3790641}. Third, deployed systems recognize actions and state transitions at runtime to drive immediate responses~\cite{10.1145/3772318.3790904, 10.1145/3772318.3791958, 10.1145/3772318.3790805}. Fourth, window-based pipelines combine evidence across successive frames to identify short-lived states and events, such as affective responses, spatial disorientation, and public-safety events~\cite{10.1145/3772318.3790745, 10.1145/3772318.3790679, 10.1145/3772318.3791432, 10.1145/3772318.3791208}.

\textbf{M4. Global sequence reasoning} (30 papers, 24.0\%): Annotation connects events from different parts of a recording to explain how a task or interaction unfolds. It considers their order and relationships, including task progress, procedural dependencies, and interaction history. First, procedural reconstruction converts instructional, workflow, or skill videos into ordered steps and their dependencies~\cite{10.1145/3772318.3790715, 10.1145/3772318.3790494, 10.1145/3772318.3790294, 10.1145/3772318.3791652, 10.1145/3772318.3790634}. Second, session-scale qualitative interpretation connects episodes across full recordings to examine how collaboration, instruction, and repair develop over time~\cite{10.1145/3772318.3791828, 10.1145/3772318.3791267, 10.1145/3772318.3791238, 10.1145/3772318.3790830, 10.1145/3772318.3791910}. Existing automated systems also construct sequence-level annotations by merging notes from short segments or carrying earlier outputs forward as context~\cite{10.1145/3772318.3790922, 10.1145/3772318.3790294, 10.1145/3772318.3790449, 10.1145/3772318.3791162}.

\subsection{Annotation Authority (D5)}
This dimension identifies whose judgment determines the annotations used substantively in the research workflow.

\textbf{A1. Researcher or domain-expert annotation} (55 papers, 44.0\%): Labels are produced through observation, interpretation, or professional judgment by researchers or domain experts. First, structured coding uses two or more independent coders and resolves disagreements through consensus. It spans gesture-elicitation studies~\cite{10.1145/3772318.3790437, 10.1145/3772318.3790491}, zoo-animal behavior analysis~\cite{10.1145/3772318.3790842, 10.1145/3772318.3791644}, and human-pet interaction analysis~\cite{10.1145/3772318.3790830}. It also supports studies of driving~\cite{10.1145/3772318.3790972}, medical training~\cite{10.1145/3772318.3791910}, human–robot interaction~\cite{10.1145/3772318.3793419}, and AI-supported design work~\cite{10.1145/3772318.3791776}. Second, interpretive studies use detailed transcription and collective discussion to explain how interactions unfold. Researchers examine selected episodes together and develop shared interpretations~\cite{10.1145/3772318.3791828, 10.1145/3772318.3791708, 10.1145/3772318.3790741, 10.1145/3772318.3790947, 10.1145/3772318.3791969}.

\textbf{A2. Participant or stakeholder annotation} (5 papers, 4.0\%): Labels are supplied by study participants or people directly connected to the activity being annotated. Their authority comes from first-person knowledge, situated experience, or collective lay judgment. First, four papers use personal knowledge to capture intentions or experiences that video alone cannot fully reveal~\cite{10.1145/3772318.3790918, 10.1145/3772318.3790596, 10.1145/3772318.3790808, 10.1145/3772318.3790836}. In Point \& Grasp, participants judged whether their gestures could plausibly grasp replacement objects, producing labels grounded in their intended actions~\cite{10.1145/3772318.3790836}. Second, aggregated participant judgments provide collective assessments of video content. Collab asked 90 participants with varied expertise to mark suspected deepfake artifacts and aggregated their annotations~\cite{10.1145/3772318.3790501}.

\textbf{A3. Automatic computational annotation} (73 papers, 58.4\%): Labels are generated by rule-based or model-based procedures without substantive human correction. Authority therefore rests with the computational pipeline. Automation turns video into behavioral measures and operational inputs at scale. Two broad uses dominate the largest category. First, empirical studies use automated tracking to measure hand and facial movement~\cite{10.1145/3772318.3791184, 10.1145/3772318.3791927}, facial-expression models to quantify affect~\cite{10.1145/3772318.3791208, 10.1145/3772318.3790745}, and gaze or head motion to derive interaction measures~\cite{10.1145/3772318.3791114, 10.1145/3772318.3791820}. Second, interactive systems consume detections and descriptions directly to provide feedback or assistance, especially in accessibility applications~\cite{10.1145/3772318.3791308, 10.1145/3772318.3790449, 10.1145/3772318.3791322, 10.1145/3772318.3790589, 10.1145/3772318.3791659, 10.1145/3772318.3790817}. More recently, general-purpose VLMs have emerged as a new source of automated video annotation in HCI. Researchers have begun using them to score street scenes~\cite{10.1145/3772318.3791292}, reconstruct software workflows~\cite{10.1145/3772318.3790294}, support gaze-based social reasoning~\cite{10.1145/3772318.3790922}, analyze video content for media authoring~\cite{10.1145/3772318.3791572}, and characterize recommendation feeds~\cite{10.1145/3772318.3790311}.

\textbf{A4. Human–machine hybrid annotation} (15 papers, 12.0\%): Labels are produced through workflows that combine machine-generated outputs with substantive human review. Two configurations recur. First, in machine-first workflows, models propose candidate events or semantic labels that researchers confirm, revise, or reject~\cite{10.1145/3772318.3791605, 10.1145/3772318.3791652, 10.1145/3772318.3791267, 10.1145/3772318.3790283, 10.1145/3772318.3791220}. Second, in human-seeded workflows, users mark a region or target and the model propagates it across the clip~\cite{10.1145/3772318.3791162, 10.1145/3772318.3790269, 10.1145/3772318.3790715, 10.1145/3772318.3790536}. Across both configurations, machines primarily support detection and tracking, while humans determine categories and exclusions. For example, \textit{Touch with Meaning} combines computer-vision models with Gemini to identify candidate social-touch intervals and label 46 contextual features. Human coders then verify and revise each label~\cite{10.1145/3772318.3791605}. Notably, general-purpose VLMs have also begun to serve as proposal generators in several hybrid workflows~\cite{10.1145/3772318.3791605, 10.1145/3772318.3791652, 10.1145/3772318.3791267, 10.1145/3772318.3790715, 10.1145/3772318.3791162}.

\section{Constructing a Taxonomy-Grounded Evaluation Benchmark}\label{sec:benchmark}
To answer when VLMs are effective across recurring video annotation tasks and how they should be used in annotation workflows (RQ2), we focus on the CHI 2026 papers identified in Section~\ref{sec:taxonomy} whose video annotation workflows involve substantial human input or use general-purpose vision-language models (VLMs). We derive five recurring task families from these workflows. We then instantiate each family with three representative tasks drawn from open datasets and conduct a comparative evaluation of three annotation workflows: VLM alone, human alone, and VLM-assisted human verification, measuring annotation quality, human effort, and monetary cost.

\subsection{Deriving Task Families from the Coded Papers}\label{sec:families}
We focus on annotation tasks whose quality still depends on human judgment, as well as tasks that researchers have recently delegated to general-purpose VLMs. We therefore filtered the 125 papers by annotation authority (D5), retaining those with at least one substantive annotation stream produced by researchers or domain experts (A1), participants (A2), or a human-machine hybrid workflow (A4). We also retained automatically generated streams (A3) when they used off-the-shelf VLMs~\cite{10.1145/3772318.3791292,10.1145/3772318.3790311,10.1145/3772318.3790294,10.1145/3772318.3790922}. We excluded streams handled entirely by conventional task-specific models, such as pose estimators and facial-expression classifiers, because such models represent established forms of automated processing rather than annotation work that still requires human judgment. This filtering yielded 74 papers.

Prior HCI reviews have used structured coding to identify recurring methodological patterns~\cite{stefanidi2023literature,pang2025understanding}. Quantitative clustering has also been used to map higher-level structure across HCI literature~\cite{liu2014chi}. We combine these approaches to identify the main forms of video annotation that still involve human judgment and to construct a systematic evaluation grounded in current HCI practice.

\paragraph{Paper profiles.} We represented each paper as a profile based on its codes across the five taxonomy dimensions. Because a paper could receive multiple codes within a dimension, each dimension was represented as a set.

\paragraph{Distance.} We measured dissimilarity between papers $i$ and $j$ using a dimension-normalized Jaccard distance~\cite{kosub2019note}:
\[
d(i,j) = 1 - \frac{1}{5}\sum_{m=1}^{5} \frac{|X_{im}\cap X_{jm}|}{|X_{im}\cup X_{jm}|},
\]
where $X_{im}$ denotes the set of codes assigned to paper $i$ in dimension $m$.

\paragraph{Clustering.} We applied $k$-medoids clustering to the distance matrix to identify recurring groups of papers with similar taxonomy profiles~\cite{kaufman2009finding}. 
We evaluated solutions with $k=3$--$7$ based on silhouette scores~\cite{rousseeuw1987silhouettes}\footnote{A higher silhouette score indicates that papers are more similar to others in their assigned cluster and more distinct from papers in neighboring clusters.} and cluster stability~\cite{hennig2007cluster}.
We finally selected $k=5$ as the most interpretable solution: its mean silhouette score was 14.5\% higher than that of $k=4$, while smaller solutions tended to merge distinct practices. In contrast, cluster stability deteriorated as $k$ increased beyond five, with worst-cluster stability decreasing by 11.4\% at $k=6$ and 28.9\% at $k=7$, indicating that additional clusters fragmented similar practices into near-duplicate groups. We therefore examined the shared characteristics of the five clusters and used them to define the task families reported in Section~\ref{sec:tasks}.
We refer to each cluster as a ``task family,'' meaning a recurring annotation configuration defined primarily by what is annotated and what reasoning the annotation requires.

\begin{table*}[h]
\centering
\caption{Five recurring annotation families and 15 representative tasks. Full definitions are provided in the supplementary materials.}
\Description{Summary of five task families, 15 benchmark tasks, task descriptions, and source datasets. The families cover long-horizon interaction interpretation, egocentric spatial grounding, interface workflow understanding, fine-grained embodied behavior recognition, and communicative signal interpretation, with three representative tasks in each family.}
\label{tab:task-families}
\small
\setlength{\tabcolsep}{4pt}
\renewcommand{\arraystretch}{1.15}
\begin{tabular}{
    @{}
    >{\raggedright\arraybackslash}p{0.15\textwidth}
    >{\raggedright\arraybackslash}p{0.27\textwidth}
    >{\raggedright\arraybackslash}p{0.45\textwidth}
    >{\raggedright\arraybackslash}p{0.13\textwidth}
    @{}
}
\toprule
\textbf{Task family} &
\textbf{Task} &
\textbf{Description} &
\textbf{Dataset} \\
\midrule

\multirow[t]{3}{0.15\textwidth}{
    \textbf{F1. Long-Horizon Interaction Interpretation}\newline
    ($n=16$)
}
& \textbf{F1-T1.}\newline Interactional dominance
& Rank four meeting participants by interactional dominance based on speaking time, interruptions, and turn control.
& AMI~\cite{carletta2007unleashing,aran2010fusing} \\
\cmidrule(lr){2-4}

& \textbf{F1-T2.}\newline Human--robot role inference
& Classify the human's role in a complete interaction based on how initiative and responsibility develop.
& HABIT~\cite{song2026habit} \\
\cmidrule(lr){2-4}

& \textbf{F1-T3.}\newline Mistake handling
& Link each marked mistake in the equipment assembly to subsequent actions and classify it as independently corrected, corrected after intervention, or unresolved.
& HoloAssist~\cite{wang2023holoassist} \\
\midrule

\multirow[t]{3}{0.15\textwidth}{
    \textbf{F2. Egocentric Spatial Grounding}\newline
    ($n=14$)
}
& \textbf{F2-T1.} \newline Visual query with temporal memory
& Identify the last time point when the queried object was visible in the egocentric video and draw its bounding box.
& Ego4D VQ2D~\cite{grauman2022ego4d} \\
\cmidrule(lr){2-4}

& \textbf{F2-T2.} \newline Manipulated-object grounding
& Draw a bounding box around the object undergoing a state change and identify it using the supplied codebook.
& Ego4D FHO~\cite{grauman2022ego4d} \\
\cmidrule(lr){2-4}

& \textbf{F2-T3.}  \newline Navigation-relevant grounding
& Locate the nearest instance of a specified road feature and report its direction relative to the viewer.
& SANPO-Real~\cite{waghmare2025sanpo} \\
\midrule

\multirow[t]{3}{0.15\textwidth}{
    \textbf{F3. Interface Workflow Understanding}\newline
    ($n=10$)
}
& \textbf{F3-T1.}\newline Local interface actions
& Timestamp observable interface interactions such as clicks and scrolling.
& GUI-World~\cite{chen2025gui} \\
\cmidrule(lr){2-4}

& \textbf{F3-T2.}\newline Workflow reconstruction
& Select the steps that occurred from a shuffled candidate list and place them in order.
& GUI-World~\cite{chen2025gui} \\
\cmidrule(lr){2-4}

& \textbf{F3-T3.}\newline Outcome and breakdown coding
& Judge whether the task was completed and localize the first breakdown when one occurred.
& GUI-World~\cite{chen2025gui} \\
\midrule

\multirow[t]{3}{0.15\textwidth}{
    \textbf{F4. Fine-Grained Embodied Behavior Recognition}\newline
    ($n=22$)
}
& \textbf{F4-T1.}\newline Behavior interval coding
& Mark the start and end times of every occurrence of 11 daily activities in home video.
& Charades~\cite{sigurdsson2016hollywood} \\
\cmidrule(lr){2-4}

& \textbf{F4-T2.}\newline Animal behavior coding
& Assign one of 12 predefined behavior codes to each clip.
& MammalNet~\cite{chen2023mammalnet} \\
\cmidrule(lr){2-4}

& \textbf{F4-T3.}\newline Attentional orientation coding
& Classify the attention state of a marked student from visible behavioral cues.
& SAV~\cite{tan2025towards} \\
\midrule

\multirow[t]{3}{0.15\textwidth}{
    \textbf{F5. Communicative Signal Interpretation}\newline
    ($n=12$)
}
& \textbf{F5-T1.}\newline Isolated sign recognition
& Identify the English gloss of an isolated American Sign Language (ASL) sign.
& WLASL~\cite{li2020word} \\
\cmidrule(lr){2-4}

& \textbf{F5-T2.} \newline Co-speech gesture interpretation
& Classify a gesture by how it conveys meaning alongside speech.
& GESRes~\cite{hensel2025richly} \\
\cmidrule(lr){2-4}

& \textbf{F5-T3.}\newline Communicative intent coding
& Identify the speaker's communicative intent, such as praise or opposition.
& MIntRec~\cite{zhang2022mintrec} \\
\bottomrule
\end{tabular}
\end{table*}

\subsection{Selecting Evaluation Tasks}\label{sec:tasks}
\begin{figure}
    \centering
    \includegraphics[width=\linewidth]{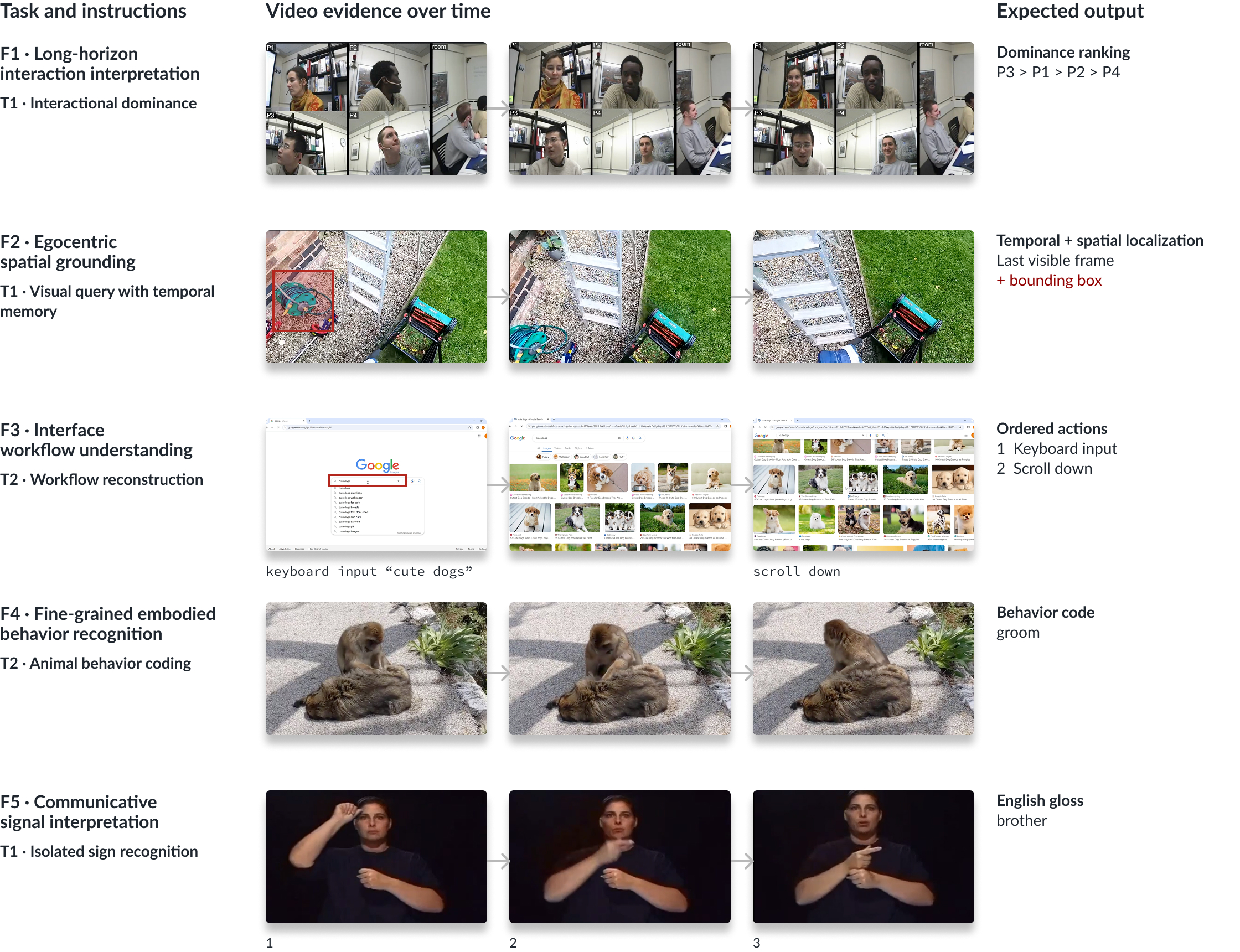}
    \caption{
    Representative examples from the five task families. Each row shows one task, the video evidence used to solve it, and the corresponding expected annotation output.
    }
    \label{fig:tasks-overview}
    \Description{Representative examples from the five task families. Each row shows one task, the video evidence used to solve it, and the corresponding expected annotation output.}
\end{figure}

Together, the five families characterize prominent video annotation practices across the 74 CHI 2026 papers. For each family, we selected three tasks that reflect its dominant phenomenon and reasoning profile while capturing meaningful within-family variation, prioritizing tasks that closely match the annotation workflows in our review. Each task uses an open dataset with reference labels, enabling direct comparison between VLM and human annotations. We denote the families as F1 through F5 and the tasks within each family as T1 through T3. Table~\ref{tab:task-families} summarizes the tasks and source datasets, while Figure~\ref{fig:tasks-overview} provides a visual overview using one representative task from each family. Figures~\ref{fig:f1_tasks}--Figures~\ref{fig:f5_tasks} in Appendix~\ref{app:agreement} expand this overview by showing all three tasks in each family, together with their instructions, representative video evidence, and expected annotation outputs. 
Supplementary materials provide the complete annotation instructions, dataset details, and output schemas.

\paragraph{F1. Long-Horizon Interaction Interpretation ($n = 16$).} This family integrates evidence from separated moments into a global account of how an interaction develops across a complete recording. It reflects CHI 2026 workflows that code participation, role dynamics, and coordination at the session level in collaborative interaction~\cite{10.1145/3772318.3791122,10.1145/3772318.3791399}, trace how roles evolve over the course of human-robot interaction~\cite{10.1145/3772318.3791828,10.1145/3772318.3791910}, and examine how assistance or intervention shapes subsequent actions~\cite{10.1145/3772318.3791708}.

\paragraph{F2. Egocentric Spatial Grounding ($n = 14$).} This family grounds physical objects and environmental features in egocentric video captured by participant-worn cameras. It reflects systems that maintain first-person visual context to retrieve objects for situated assistance~\cite{10.1145/3772318.3791059}, identify manipulated objects and hand-object interaction~\cite{10.1145/3772318.3790817,10.1145/3772318.3791308}, and ground landmarks and salient environmental features for accessible navigation~\cite{10.1145/3772318.3790589,10.1145/3772318.3790604}.

\paragraph{F3. Interface Workflow Understanding ($n = 10$).} This family codes screen-recorded activity at both event and workflow scales, from individual interface actions to ordered task steps and task completion. It reflects workflows that infer discrete interface gestures and local state changes~\cite{10.1145/3772318.3790283,10.1145/3772318.3790294}, reconstruct complete digital workflows at the sequence level~\cite{10.1145/3772318.3790294,10.1145/3772318.3791776}, and analyze usability outcomes during multi-step interface use~\cite{10.1145/3772318.3791293,10.1145/3772318.3791399}.

\paragraph{F4. Fine-Grained Embodied Behavior Recognition ($n = 22$).} This family applies study-specific codebooks to discrete embodied behaviors within bounded clips, covering body movement, object manipulation, and attention. It reflects study-specific coding of temporally bounded human actions~\cite{10.1145/3772318.3790437,10.1145/3772318.3793419}, ethogram-based coding in animal-computer interaction research~\cite{10.1145/3772318.3790842,10.1145/3772318.3791644}, and inference of attention or engagement from visible cues such as gaze~\cite{10.1145/3772318.3790972,10.1145/3772318.3791481}.

\paragraph{F5. Communicative Signal Interpretation ($n = 12$).} This family interprets visible behavior as a communicative signal whose meaning depends on social convention and interactional context. It reflects systems that recognize signed movements for accessibility~\cite{10.1145/3772318.3790774}, workflows that interpret how gesture and speech jointly convey meaning~\cite{10.1145/3772318.3790641,10.1145/3772318.3790491}, and workflows that infer intended social meanings from multimodal behavior in context~\cite{10.1145/3772318.3790944,10.1145/3772318.3791273}.

\subsection{Evaluation Design}\label{sec:design}
We used the 15 tasks described in Section~\ref{sec:tasks} to construct an evaluation benchmark and compared three workflows with different allocations of annotation authority.

\subsubsection{Benchmark Construction}
We drew evaluation videos from the source datasets introduced in Section~\ref{sec:tasks}. Each task contained 100 videos, except F1-T1, which contained 59. Each task instance combined a video with all information needed for annotation. Depending on the task, this included an object query, transcript, codebook, candidate step list, or marked target. The resulting benchmark comprised 1,459 task instances. 
The code, prompts, task specifications, and annotation interface are available in a
\href{https://anonymous.4open.science/r/vlm-annotation-chi27submission/}{\underline{linked anonymous repository}}.

\begin{figure}
    \centering
    \includegraphics[width=\linewidth]{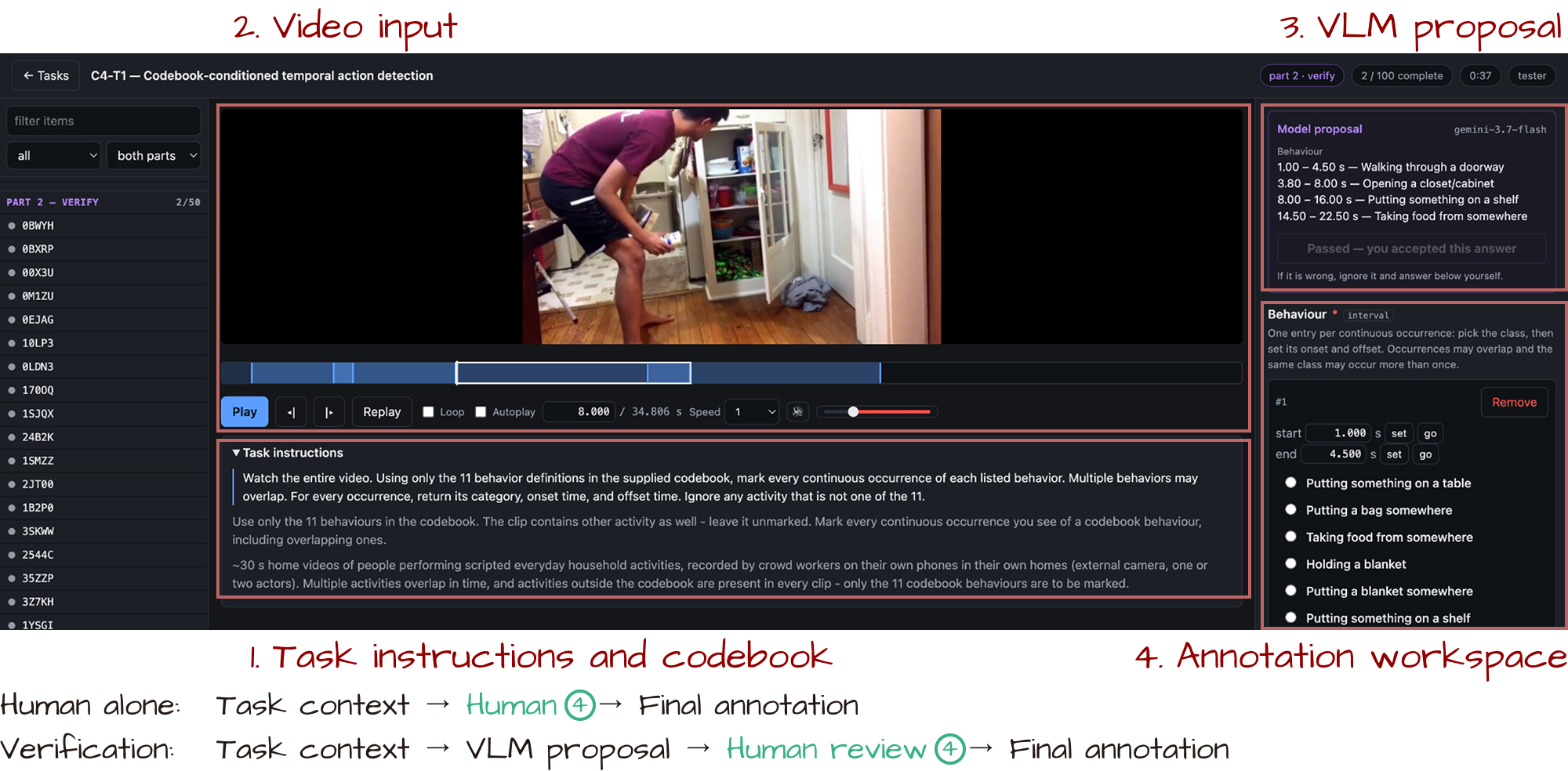}
    \caption{
    Annotation interface and annotation authority conditions. Annotators viewed the task instructions and codebook, video input, and annotation workspace. In the human-alone condition, humans produced the annotation directly from the task context; in the verification condition, humans reviewed the VLM proposal and produced the final annotation.  
    }
    \label{fig:user_interface}
    \Description{Annotation interface and annotation authority conditions. Annotators viewed the task instructions and codebook, video input, and annotation workspace. In the human alone condition, humans produced the annotation directly from the task context; in the verification condition, humans reviewed the VLM proposal and produced the final annotation.}
\end{figure}

\subsubsection{Annotation Workflow Conditions}
We evaluated three annotation-authority conditions: VLM alone, human alone, and VLM with human verification. These conditions assigned responsibility for the final annotation to the model, human annotators, or humans reviewing and correcting model proposals, respectively. Across conditions, humans and the VLM received the same context information. The verification condition additionally presented the annotation label generated by the VLM to the human annotator. Figure~\ref{fig:user_interface} illustrates the annotation interface and the information presented under the human alone and verification conditions.

\paragraph{VLM alone.} We used \texttt{gemini-3.7-flash}~\cite{team2023gemini} as the general-purpose VLM for annotation. We selected it based on pilot testing because it was the strongest-performing model available to us that accepted video files, including their audio tracks, directly through its API. 
At the time of model selection, other leading models we considered, including GPT models, did not support native video input and would have required preprocessing videos into sampled frames and separate audio files, while smaller or earlier models with native video support performed substantially worse. 
These properties made \texttt{gemini-3.7-flash} a practical representative of efficient API-based VLMs for large-scale video annotation, where annotation quality must be balanced against throughput and inference cost~\cite{googledeepmind2026gemini37flash}. Each request contained the complete task input, the applicable label space, and a JSON schema corresponding to the reference annotation format.

\paragraph{Human alone.} For each task, we randomly divided the video instances evenly between the human-alone and verification conditions. Tasks containing 100 instances contributed 50 to each condition, with no overlap between the two sets. F1-T1, which contained 59 videos, was divided between conditions as evenly as possible. Two researchers independently annotated the human-alone set through a custom web interface (see Figure~\ref{fig:user_interface}). The interface displayed the task instructions and codebook beside a video player. It supported categorical labels, rankings, ordering steps, temporal intervals, and bounding boxes. Annotation time was recorded for each video. 

\paragraph{VLM with human verification.} The same researchers annotated the complementary half of each task by verifying the VLM outputs. The interface was prefilled with the model annotation. Annotators could accept the labels or revise any field. The interface retained both the original labels and the final annotation. It also recorded each accepted or edited decision and the active verification time. 

\subsubsection{Evaluation Measures}
We evaluated the three conditions in terms of annotation quality, time, and monetary cost. We computed each researcher’s results and then averaged them. 
We additionally assessed pairwise annotation consistency using each task's primary evaluation metric.

\paragraph{Annotation quality.} We chose the evaluation metric according to what annotators were asked to produce. Rankings were scored using Kendall's $\tau_b$~\cite{kendall1945treatment}. Classifications used macro-F1. Temporal annotations used event-level F1 under task-specific tolerances. Localization was evaluated with temporal-window and intersection-over-union (IoU) thresholds.
Ordered workflows used normalized edit similarity. Codebook-based word selection was evaluated using top-1 accuracy.
We applied the same scoring implementation across all three conditions. The supplementary materials report the complete raw scores.
Because the tasks used different evaluation metrics, we normalized the two VLM-based conditions relative to human-alone performance using the Human-Normalized Score (HNS)~\cite{volodymyr2015human,srivastava2022beyond}, an established measure of machine performance relative to human performance. On this scale, 0 represents chance performance, 100 represents mean human-alone performance, and values above 100 indicate performance exceeding the human reference.

For workflow \(w\) on task \(t\), we calculated:
\[
\mathrm{HNS}_{w,t} = 100 \times \frac{S_{\mathrm{w},t}-S_{\mathrm{chance},t}}{S_{\mathrm{human},t}-S_{\mathrm{chance},t}},
\]
where $S_{\mathrm{human},t}$ is the mean score of the two human-alone annotators and $S_{\mathrm{chance},t}$ is the corresponding naive chance baseline. For each task, VLM-alone scores were computed on subset \(A\) and compared with human-alone scores on the same instances, whereas verification scores were computed on the complementary subset \(V\).

\paragraph{Human time.} We compared the annotation time between human-alone and VLM with human verification conditions.

\paragraph{Monetary cost.} We calculated VLM cost from measured input and output token usage. The published prices for \texttt{gemini-3.7-flash} were \$0.75 per million input tokens and \$3.75 per million output tokens. Human cost was estimated by multiplying annotation time by the local compensation rate.

\section{Results}\label{sec:results}
Because the benchmark comprises 15 purposively selected tasks, the analyses below are descriptive, task-level comparisons. We therefore assess whether the observed patterns are consistent with each hypothesis rather than treating them as population-level tests across HCI video annotation tasks.
We organize the results around six research hypotheses. The first three examine whether VLM performance varies across recurring dimensions of HCI video annotation identified in Section~\ref{sec:taxonomy}.
\begin{description}
    \item[\textbf{RH1 (Viewpoint).}]
    VLM annotation performance differs systematically across video viewpoints.

    \item[\textbf{RH2 (Phenomenon).}]
    VLM annotation performance differs systematically across the phenomena being annotated.

    \item[\textbf{RH3 (Reasoning).}]
    VLM annotation performance differs systematically across reasoning requirements, with longer-horizon reasoning expected to be more difficult.
\end{description}

The remaining three hypotheses compare the annotation quality, human time, and monetary cost of the three workflow conditions.
\begin{description}
    \item[\textbf{RH4 (Accuracy).}]
    Human verification of VLM output yields more accurate annotations than either VLM-alone or human-alone annotation.
    \item[\textbf{RH5 (Human time).}]
    Human verification of VLM output requires less human time than annotation from scratch.
    \item[\textbf{RH6 (Monetary cost).}]
    VLM-alone annotation and VLM with human verification both have substantially lower monetary costs than human-alone annotation.
\end{description}

\subsection{Overall Annotation Accuracy}\label{sec:accuracy}
\begin{figure}
    \centering
    \includegraphics[width=\linewidth]{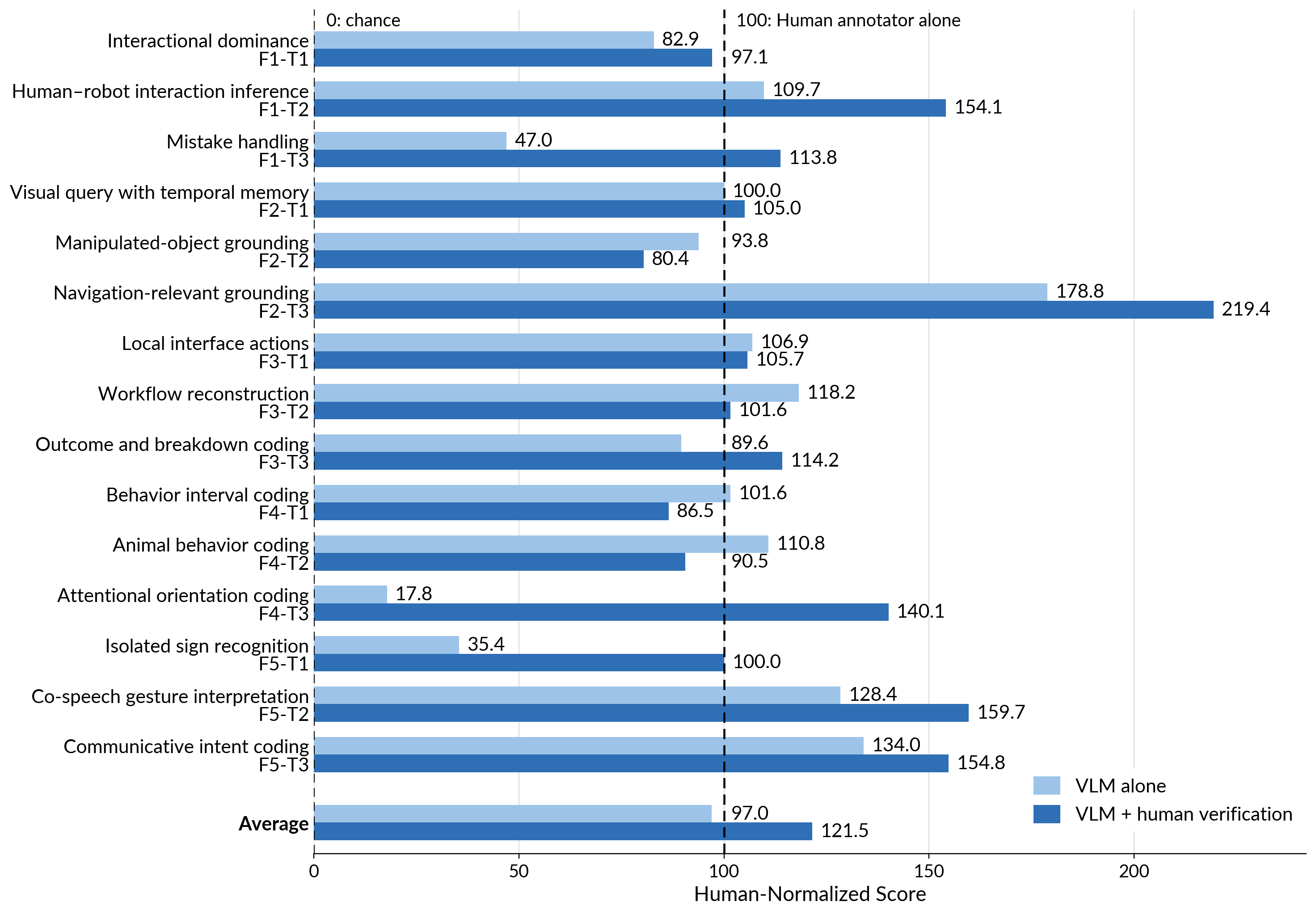}
    \caption{Annotation accuracy across the 15 benchmark tasks. Human-normalized scores set chance performance to 0 and human-alone performance to 100; VLM-alone annotation was comparable to human-alone annotation on average (mean HNS = 97.0), while human verification achieved the highest mean accuracy (HNS = 121.5).}
    \Description{Horizontal grouped bar chart comparing VLM-alone annotation with VLM plus human verification on 15 tasks using Human-Normalized Score, where 0 represents chance and 100 represents human-alone performance. Mean scores are 97.0 for VLM alone and 121.5 for verification. Verification produces especially large gains for attentional orientation coding, from 17.8 to 140.1; mistake handling, from 47.0 to 113.8; and isolated sign recognition, from 35.4 to 100.0. Navigation-relevant grounding has the highest scores, at 178.8 and 219.4.}
    \label{fig:hns}
\end{figure}

Figure~\ref{fig:hns} reports HNS across the 15 tasks. Averaged equally across tasks, VLM-alone annotation had a mean HNS of 97.0, close to the human-alone reference of 100.0. The verification condition had the highest mean HNS at 121.5. The VLM alone reached or exceeded human-alone performance on 9 of 15 tasks ($\mathrm{HNS}=100.0$--$178.8$). Human verification reached or exceeded human-alone performance on 11 of 15 tasks and outperformed VLM-alone annotation on 10 tasks.

\subsection{Performance by Annotation Dimension}\label{sec:dimensions}
We examine how VLM-alone accuracy varied across video viewpoint (RH1), annotated phenomenon (RH2), and reasoning requirement (RH3).

\subsubsection{Video Viewpoint}\label{sec:viewpoint}
VLM annotation accuracy did not differ systematically across video viewpoints. Both external scene and egocentric recordings included tasks ranging from well below to above human-alone performance. For external scene views (V1), VLM-alone annotation accuracy ranged from attentional orientation coding (F4-T3, $\mathrm{HNS}=17.8$) to communicative intent coding (F5-T3, $\mathrm{HNS}=134.0$). Egocentric video (V2) showed similarly broad variation. Mistake handling (F1-T3) achieved an $\mathrm{HNS}$ of 47.0, whereas the three tasks in the Egocentric Spatial Grounding family ranged from 93.8 to 178.8. Screen recordings (V4) showed promising and comparatively consistent VLM-alone annotation accuracy: local interface actions (F3-T1), workflow reconstruction (F3-T2), and outcome and breakdown coding (F3-T3) fell within a relatively narrow range of $\mathrm{HNS}=89.6$--$118.2$.
Within external scene views, communicative intent coding drew on multiple conversational cues and achieved an $\mathrm{HNS}$ of 134.0 (F5-T3). In contrast, attentional orientation coding depended on subtle head and gaze changes by a target student within a wider classroom view and achieved an $\mathrm{HNS}$ of 17.8 (F4-T3).

Within this benchmark, the viewpoint-level pattern predicted by RH1 was not observed. Performance varied substantially within viewpoint categories.

\subsubsection{Annotated Phenomenon}\label{sec:phenomenon}
VLM annotation accuracy differed substantially across annotated phenomena. Tasks involving directly observable movement and interaction (P1, P6, and P7) or interface content (P3) generally reached or exceeded human-alone accuracy. In contrast, tasks requiring judgments of attention or mental states (P2) performed substantially worse. The VLM exceeded human-alone annotation on behavior interval coding (F4-T1, $\mathrm{HNS}=101.6$), animal behavior coding (F4-T2, $\mathrm{HNS}=110.8$), local interface actions (F3-T1, $\mathrm{HNS}=106.9$), and workflow reconstruction (F3-T2, $\mathrm{HNS}=118.2$). Attention and affect (P2) presented a markedly greater challenge. Attentional orientation coding produced the lowest VLM-alone score (F4-T3, $\mathrm{HNS}=17.8$). Interactional dominance also remained below human-alone performance (F1-T1, $\mathrm{HNS}=82.9$).
Within human-human communication (P5), the VLM exceeded human-alone performance on co-speech gesture interpretation (F5-T2, $\mathrm{HNS}=128.4$) and communicative intent coding (F5-T3, $\mathrm{HNS}=134.0$). By contrast, the VLM performed poorly on isolated sign recognition (F5-T1, $\mathrm{HNS}=35.4$), which required a precise mapping between a conventional sign form and a domain-specific gloss.

RH2 received mixed descriptive support. Some contrasts aligned with annotated phenomenon, but performance also varied substantially within categories.

\subsubsection{Reasoning Requirements}\label{sec:reasoning}
VLM annotation accuracy did not decline systematically as reasoning extended from local events to longer temporal sequences. Both local temporal reasoning (M3) and global sequence reasoning (M4) included tasks with strong and weak performance. Global-sequence tasks ranged from mistake handling (F1-T3, $\mathrm{HNS}=47.0$) to workflow reconstruction (F3-T2, $\mathrm{HNS}=118.2$), while local-temporal tasks ranged from attentional orientation coding (F4-T3, $\mathrm{HNS}=17.8$) to communicative intent coding (F5-T3, $\mathrm{HNS}=134.0$).
Thus, the observed scores did not follow the predicted decline from local to global reasoning.

Within global sequence reasoning (M4), the VLM exceeded human-alone performance on workflow reconstruction (F3-T2, $\mathrm{HNS}=118.2$) but underperformed on outcome and breakdown coding (F3-T3, $\mathrm{HNS}=89.6$) and mistake handling (F1-T3, $\mathrm{HNS}=47.0$). In both tasks, it incorrectly treated subsequent activity as evidence of successful resolution, interpreting continued instruction as mistake correction and recovery sequences as completion without breakdown.

For local temporal reasoning (M3), the VLM accurately localized clearly defined and observable events. It reached or exceeded human-alone performance when locating human behavior intervals (F4-T1, $\mathrm{HNS}=101.6$) and timestamping screen interface events (F3-T1, $\mathrm{HNS}=106.9$).
For spatial reasoning, the VLM achieved near-human performance on manipulated-object grounding but remained less precise than human-alone annotation (F2-T2, $\mathrm{HNS}=93.8$).

RH3 was partially supported. Performance did vary across reasoning requirements, but it did not decline with temporal horizon as predicted.

\begin{figure}
    \centering
    \includegraphics[width=\linewidth]{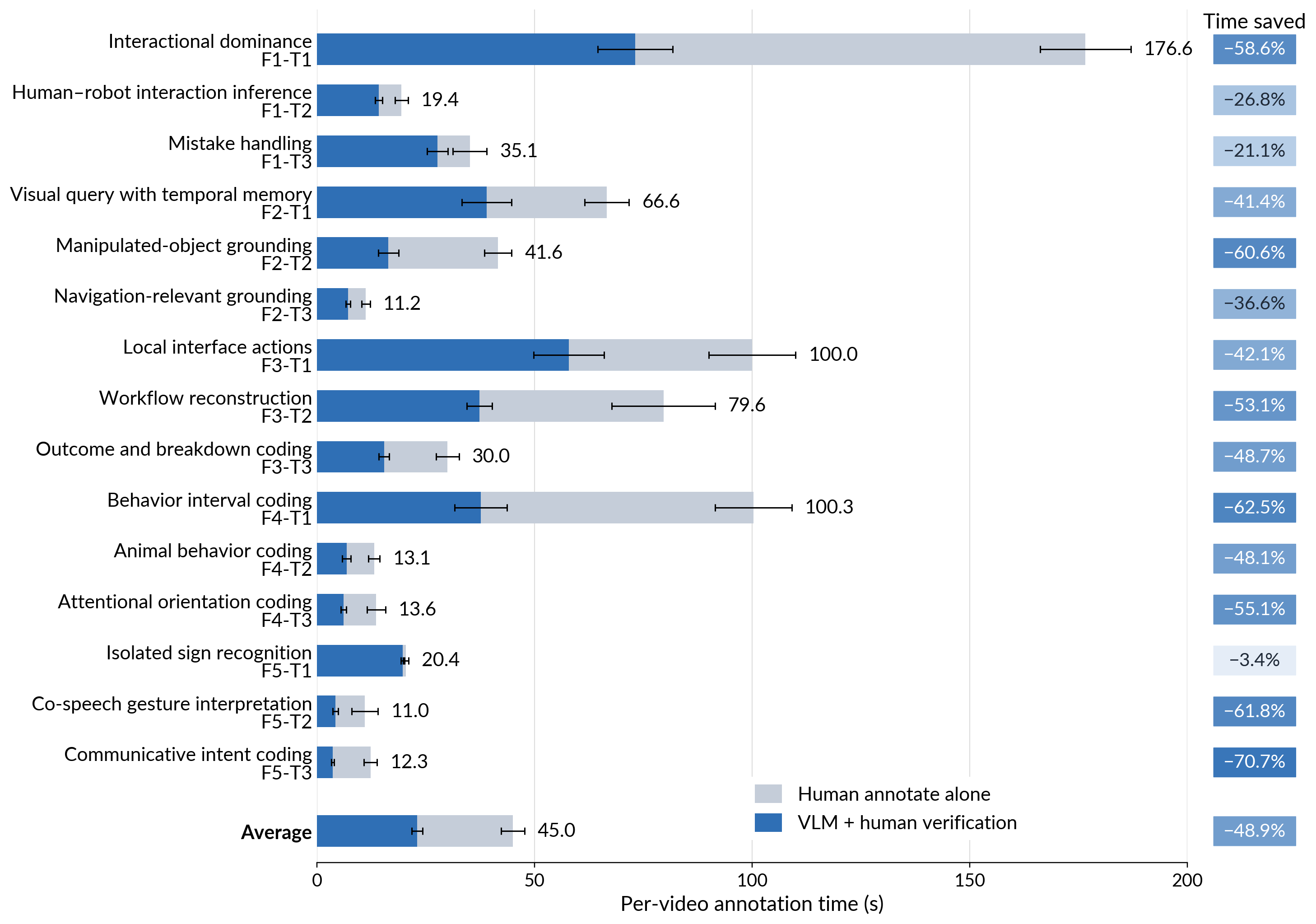}
    \caption{Human annotation time across tasks. Verifying VLM-generated annotations reduced human annotation time by 48.9\% overall, though savings varied substantially across tasks. Error bars denote standard errors.}
    \label{fig:annotationtime}
    \Description{Horizontal bar chart comparing human-alone annotation time with VLM-assisted human verification for each of the 15 tasks, with standard-error bars and the percentage saved shown at right. Verification is faster for every task and reduces the pooled mean from 45.0 to 23.0 seconds per video, a 48.9 percent reduction. Savings range from 3.4 percent for isolated sign recognition to 70.7 percent for communicative intent coding.}
\end{figure}

\subsection{Human-VLM Workflow Comparison}\label{sec:workflows}

We next compare the three workflow conditions in terms of annotation accuracy (RH4), human annotation time (RH5), and monetary cost (RH6).

\subsubsection{Accuracy of Human Verification}\label{sec:verification-accuracy}
Human verification improved annotation accuracy overall and produced the strongest aggregate workflow (mean $\mathrm{HNS}=121.5$, compared with 97.0 for VLM-alone and 100.0 for human-alone annotation). However, the effectiveness of human verification varied markedly across tasks. The largest gains occurred on the three tasks with the lowest VLM-alone scores. Attentional orientation coding (F4-T3) increased from $\mathrm{HNS}=17.8$ to 140.1, mistake handling (F1-T3) increased from 47.0 to 113.8, and isolated sign recognition (F5-T1) increased from 35.4 to 100.0.
The particularly large gain for attentional orientation coding (F4-T3) may partly result from imprecision in the bounding boxes used to identify the target person. Human verifiers could infer the intended target from the surrounding scene, whereas the VLM was more susceptible to these localization errors. We discuss this data-quality issue further in Section~\ref{sec:limitations}.

Verification was less effective when VLM output was already accurate, particularly on tasks requiring precise categories or boundaries. For behavior interval coding (F4-T1), verification reduced $\mathrm{HNS}$ from 101.6 to 86.5. For animal behavior coding (F4-T2), it reduced $\mathrm{HNS}$ from 110.8 to 90.5. Verification was also unreliable when the underlying construct was difficult for both humans and the VLM.
Notably, verification increased human-human annotation agreement on 11 of the 15 tasks. Appendix~\ref{app:agreement} reports the complete task-level results, including increases from 0.761 to 0.925 for workflow reconstruction (F3-T2) and from 0.680 to 0.760 for outcome and breakdown coding (F3-T3).

The aggregate pattern was consistent with RH4: verification had the highest mean $\mathrm{HNS}$, but its advantage was not uniform across tasks.

\subsubsection{Human Annotation Time}\label{sec:time}

As shown in Figure~\ref{fig:annotationtime}, human verification had a lower observed mean annotation time on all 15 tasks. Pooling all videos across the 15 tasks, human-alone annotation required $45.0$ seconds ($SE=2.7s$) per video, whereas human verification required $23.0$ seconds ($SE=1.2s$), a 48.9\% reduction in human annotation time. The relative reduction ranged from 3.4\% for isolated sign recognition (F5-T1) to 70.7\% for communicative intent coding (F5-T3). The largest absolute savings occurred on more time-intensive tasks that required reviewing longer interactions or marking multiple temporal intervals: interactional dominance (F1-T1) decreased from 176.6 to 73.1 seconds per video, while behavior interval coding (F4-T1) decreased from 100.3 to 37.6 seconds.

Consistent with RH5 within this benchmark, mean verification time was lower for all 15 tasks, yielding a pooled reduction of 48.9\%.

\begin{figure}[h]
    \centering
    \includegraphics[width=0.6\linewidth]{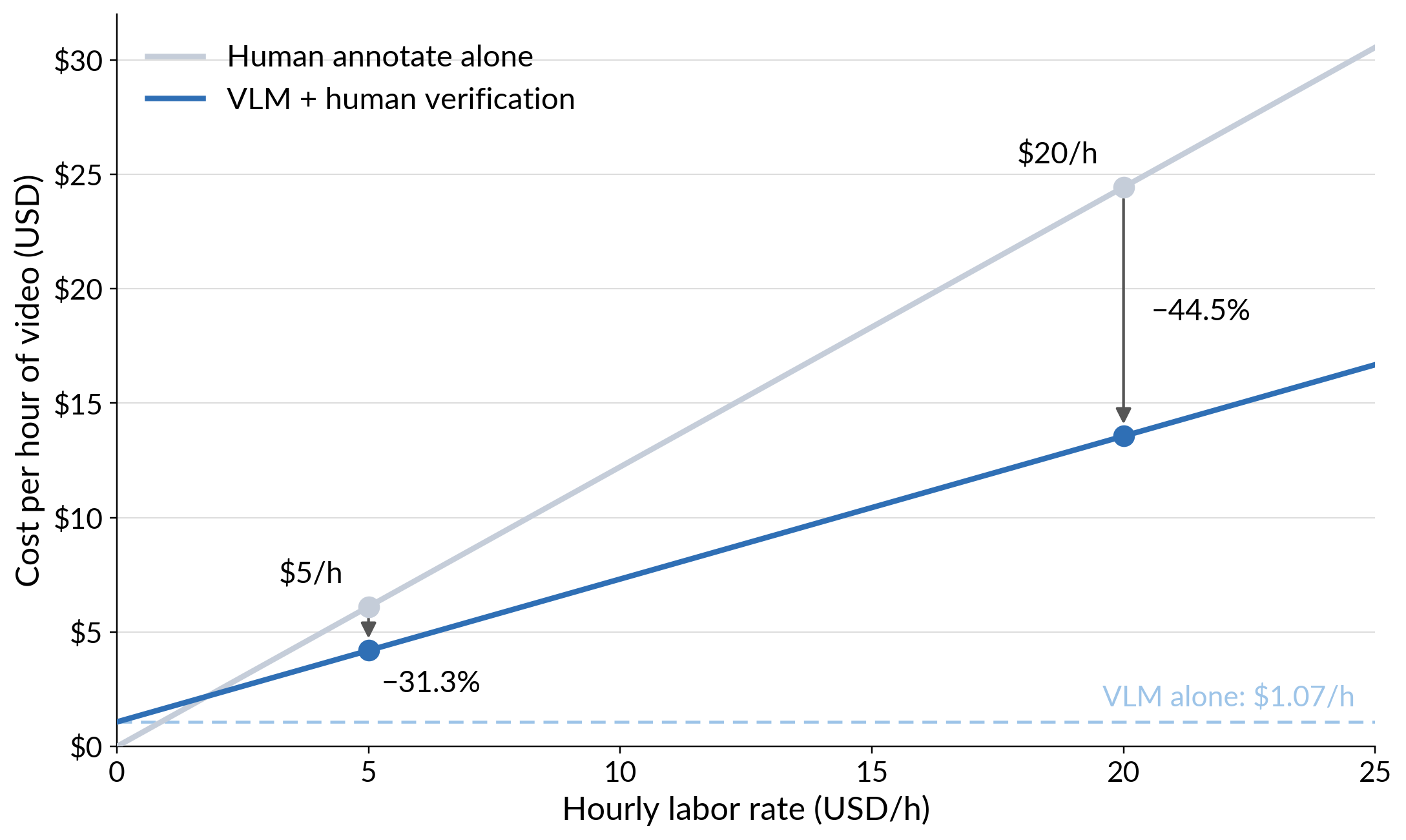}
    \caption{Estimated annotation cost per hour of video. VLM with human verification costs 31.3\% less than human-alone annotation at a \$5 hourly labor rate and 44.5\% less at \$20, while VLM-alone annotation remains fixed at \$1.07 per hour of video.}
    \Description{Line chart of estimated annotation cost per hour of video as the hourly labor rate increases from 0 to 25 dollars. Human-alone cost rises more steeply than the cost of VLM plus human verification. Verification costs 31.3 percent less at a 5-dollar hourly rate and 44.5 percent less at a 20-dollar rate, while VLM-alone annotation remains fixed at 1.07 dollars per hour of video.}
    \label{fig:annotationcost}
\end{figure}

\subsubsection{Monetary Cost}\label{sec:cost}
VLM annotation of the full benchmark cost \$16.03. The 1,459 API requests covered 895.7 minutes of video ($\approx 15$ hours) and consumed 20,770,000 input tokens and 120,000 output tokens. This expenditure corresponds to \$1.07 per hour of video. Costs differed across datasets because fixed prompt overhead accounted for a greater share of token use when datasets comprised many short clips. Appendix~\ref{app:cost} reports the complete dataset-level breakdown.
Because the human-alone and verification conditions were each evaluated on half of the benchmark, we doubled the observed times to estimate the human effort required for the full benchmark. This extrapolation yielded estimates of 18.24 hours for human-alone annotation and 9.32 hours for human verification, equivalent to 1.22 and 0.62 hours of human labor per hour of video, respectively. Figure~\ref{fig:annotationcost} converts these estimates into monetary costs across hourly labor rates, including the VLM inference cost for the verification workflow. At hourly labor rates of \$5 and \$20, the estimated cost of VLM-alone annotation was 82.4\% and 95.6\% lower than that of human-alone annotation, respectively. The corresponding reductions for VLM~+~human verification were 31.3\% and 44.5\%.

RH6 was supported. Under the evaluated task mix, model price, and labor-rate assumptions, both VLM-based workflows had lower estimated costs than human-alone annotation.

\section{Discussion}\label{sec:discussion}
We first interpret the results in terms of the capability boundaries of general-purpose VLMs and the trade-offs among human-VLM annotation workflows, and then translate these findings into actionable guidelines for researchers.

\subsection{Interpreting Task-Level Patterns in VLM Performance}\label{sec:boundaries}
Across the 15 selected tasks, no single taxonomy dimension explained the full range of VLM performance. Our benchmark was designed to compare diverse tasks, not to represent all video annotation practices in HCI. We therefore treat the patterns below as tentative explanations for differences in performance across tasks.

\paragraph{Viewpoint alone did not determine difficulty in our benchmark.}
Egocentric footage was not inherently difficult for the VLM. Across the spatial-grounding tasks, accuracy remained near or above the human-alone baseline despite motion blur, hand occlusion, and rapid viewpoint changes. Egocentric capture may therefore support VLM assistance in HCI studies using wearable and first-person cameras. More broadly, viewpoint appears to affect performance by shaping how clearly task-relevant evidence is represented. Screen recordings, for example, may make interface states and discrete visual transitions easier to distinguish than physical-world activity. The marked difference in accuracy between communicative intent and attentional orientation coding, despite their shared viewpoint, further suggests that performance depended on the clarity of task-relevant evidence rather than viewpoint alone.

\paragraph{Performance depended on the available evidence and required knowledge.}
The VLM matched or exceeded human-alone annotation when visible actions or state transitions were mapped to clearly specified categories supported by salient cues, as in behavior interval and animal behavior coding. Performance declined when labels depended on fine perceptual distinctions, latent states, specialized conventions, or task-specific knowledge, as in attentional orientation coding, sign recognition, and mistake handling.
Co-speech gesture and communicative intent tasks provided contextual information and yielded better performance than isolated sign recognition. This difference suggests that the VLM used general social and semantic knowledge more reliably than it mapped isolated signs to their correct labels. Thus, the key distinction was not perception versus interpretation, but whether the label could be determined from observable evidence and broadly available contextual knowledge.

\paragraph{Reasoning type appeared more informative than temporal horizon in the selected tasks.}
The VLM more reliably reconstructed observable events and their order than diagnosed errors or linked earlier actions to later outcomes. The contrast between workflow reconstruction and interactional dominance further suggests that extended context was manageable when it involved observable event sequences, but more difficult when it required aggregating subtle participation cues. The VLM also identified event timing and boundaries when the target event had a clear visual definition. For spatial grounding, the VLM could approximate object locations and directions, but our results do not establish the geometric precision needed for exact coordinates or tight bounding boxes.

\paragraph{Subtle failures and breakdowns were more likely to be missed.}
In both mistake handling and breakdown coding, the model identified explicit actions and successful outcomes more reliably than missing actions, failures, and other states with few visible cues. VLM-generated annotations may therefore undercount breakdowns, unresolved errors, disengagement, and other behaviors that are important to HCI research but difficult to observe directly.

\subsection{Comparing Human-VLM Annotation Workflows}\label{sec:workflow-discussion}
The results for RH4--RH6 revealed no uniformly superior workflow. VLM-alone annotation had an $\mathrm{HNS}$ close to the human-alone reference, but performance varied substantially across tasks. Human verification had the highest task-averaged $\mathrm{HNS}$ while requiring less human annotation time and incurring lower estimated costs than human-alone annotation, although its accuracy benefit was task-dependent. Task-level contrasts suggest that verification produced the largest gains when VLM-alone performance was weak but reviewers could readily recognize and correct the model's errors. It was less effective when the VLM output was already accurate, because human edits sometimes introduced incorrect labels or temporal boundaries, and when the underlying task was ambiguous for both humans and the VLM. In the latter case, reviewers who began with the same model-generated annotation could agree with one another even when the shared annotation was incorrect.

The largest observed time savings occurred for more time-intensive tasks and when reviewers could assess the proposed labels directly. Savings were smaller when reviewers effectively had to reconstruct the annotations from scratch. Under the model prices and labor rates examined, inference costs were small relative to human labor, so reductions in human time produced greater monetary savings as labor rates increased. Overall, these results suggest that verification is most useful when VLM output provides a substantive starting point that reviewers can evaluate and correct more efficiently than annotating independently. Its value should nevertheless be established through a task-specific pilot rather than assumed from aggregate performance.

\subsection{Guiding Questions for Researchers Using VLMs for Video Annotation}\label{sec:guidelines}
The following questions translate our findings into actionable guidance for deciding whether and how to use VLMs for video annotation. Organized around task definition, evidence requirements, and collaboration design, the questions below operationalize the capabilities identified in RH1--RH3 and the workflow trade-offs identified in RH4--RH6. Because our benchmark included a deliberately selected set of 15 tasks, these recommendations may not apply to every annotation setting. Researchers should first evaluate VLM performance on a sample of their own data.

\guidelinegroup{taxD2}{Can the Annotation Task Be Operationalized?}

\guidelinequestion{1}{Can the labels be defined through operational criteria that an unfamiliar annotator could apply?}
Performance was strongest when labels could be expressed through clear natural-language rules and supported by salient visual evidence. Interface action and workflow coding exemplified this profile, whereas tasks relying on implicit judgment performed poorly. The prompt should therefore function as a codebook, specifying label definitions, decision criteria, output formats, and representative examples.

\guidelinequestion{2}{Does the annotation scheme include failures or other low-salience states?}
In the two evaluated tasks that explicitly included failures or breakdowns, the VLM more often missed low-salience labels than explicit actions or successful outcomes. Researchers should therefore measure class-specific recall for such labels in their own data.

\guidelinequestion{3}{Are the video units aligned with the annotation granularity?}
Video should be segmented into units that provide sufficient context while corresponding to the intended event.

\guidelinegroup{taxD3}{What Evidence and Reasoning Does the Task Require?}

\guidelinequestion{4}{Is the relevant evidence clearly visible and salient?}
The viewpoint alone did not explain performance. Challenging egocentric footage could still support near-human accuracy when the target evidence remained identifiable. Researchers should assess target size, visibility, and occlusion directly rather than treating camera viewpoint as a proxy for task difficulty.

\guidelinequestion{5}{Does the label describe observable behavior or a state that must be inferred?}
The VLM performed more reliably on body movement, physical objects, and digital interfaces. Tasks involving attention and affect produced the weakest results. Where the research question permits, researchers can instead annotate observable indicators, such as head orientation or gaze target, and combine them during subsequent analysis. This decomposition makes both model outputs and human review more auditable.

\guidelinequestion{6}{Can the label be interpreted through general contextual knowledge, or does it require specialized conventions?}
The VLM performed well on co-speech gesture and communicative intent tasks, where context helped interpret the observed behavior. It performed poorly on isolated American Sign Language recognition, which required matching specific signs to their correct labels. Researchers should determine whether prompts, codebooks, and examples can adequately provide the required conventions. Otherwise, expert annotation or verification remains necessary.

\guidelinequestion{7}{What spatial and temporal precision does the downstream analysis require?}
Our results suggest that VLMs can support approximate object locations, directions, and event boundaries, but do not establish their reliability for exact coordinates or tight bounding boxes. Evaluation metrics and output formats should therefore reflect the precision required by the intended analysis.

\guidelinegroup{taxD5}{Which Human-AI Collaboration Mode Is Most Appropriate?}

\guidelinequestion{8}{Has a representative subsample been annotated independently by humans?}
Aggregate comparability with human-alone annotation concealed substantial variation across tasks. Before scaling, researchers should evaluate accuracy and class-specific failure patterns on a representative subsample whose reference labels were produced without exposure to VLM output. Agreement among reviewers who verify the same VLM predictions should not be treated as independent evidence of reliability, because the shared predictions can induce correlated errors.

\guidelinequestion{9}{Can reviewers recognize and correct VLM errors more efficiently than producing labels from scratch?}
The value of verification depended on whether errors were visible and readily correctable, rather than on VLM accuracy alone. Low standalone performance does not rule out verification if the VLM output narrows the decision, allowing reviewers to reliably repair its errors. Conversely, verification saves little when reviewers must reconstruct the label independently. Researchers should therefore compare annotation and verification time directly in specific tasks.

\guidelinequestion{10}{When VLM accuracy is sufficiently high, is video-level verification still necessary?}
Verification did not consistently improve strong VLM outputs and occasionally introduced inconsistencies in categorical labels or temporal boundaries. When task-specific evaluation demonstrates high accuracy across relevant classes, VLM-alone annotation with representative audits may be preferable to exhaustive review. Pilot results can be used to target review toward rare or low-salience categories.

\guidelinequestion{11}{Do both humans and the VLM have difficulty with the task?}
A plausible VLM output can increase agreement by anchoring reviewers on the same interpretation without increasing accuracy. Higher agreement among reviewers who begin from the same output should therefore not be interpreted as evidence of validity. When independent human annotations also show weak performance, researchers should clarify or decompose the annotation scheme before optimizing the workflow.

\guidelinequestion{12}{Do the end-to-end savings hold at the intended scale?}
Time and monetary savings depend on baseline annotation time, the effort required to inspect and correct VLM outputs, corpus size, labor rates, and inference costs. Researchers should estimate these quantities from a task-specific pilot. In our benchmark, inference costs were small relative to human labor, allowing modest per-video time reductions to accumulate into substantial savings at scale. However, datasets containing many short clips incurred higher per-minute inference costs because of fixed prompt overhead.

\section{Limitations and Future Work}\label{sec:limitations}
First, our review covers a bounded sample of recent HCI research. We screened the titles, keywords, abstracts, and introductions of CHI 2026 full papers using video- and annotation-related terms, potentially missing papers that used other terminology or described video coding elsewhere. A single year of CHI also cannot represent practices across specialized HCI venues or adjacent fields. Therefore, our taxonomy characterizes recurring practices in this corpus rather than exhaustively mapping video annotation. Future reviews could extend the analysis across venues and years using broader full-text searches.

Second, taxonomy development was interpretive and researcher-led. Krippendorff's $\alpha$ above .80 across all dimensions indicates consistent coding but does not ensure that every meaningful distinction was captured, and the taxonomy reflects the perspectives of its three coders. Future work could test and refine it through broader expert review. We did not use computational assistance in the final process because our exploratory attempts did not yield useful conceptual distinctions. Future research could examine how LLMs might support taxonomy development at larger scales while preserving human interpretive judgment.

Third, the benchmark covers five task families without fully representing their diversity. Although the families were systematically derived from the taxonomy profiles of 74 papers, selecting three tasks per family was purposive and constrained by available open video datasets, potentially underrepresenting some annotation practices. Because taxonomy dimensions and other task characteristics co-varied across datasets, cross-task comparisons identify associations rather than the causal effects of viewpoint, phenomenon, or reasoning requirement. The resulting capability patterns and guiding questions should therefore be treated as empirically grounded observations rather than universal decision rules.

Fourth, the benchmark is modest in scale and partly limited by source-provided reference annotations. Although it contains 1,459 video instances, the relevant unit for cross-task generalization is the task ($n$=15). These tasks were purposively selected under open-data constraints rather than sampled from a defined population, and the number of instances provides limited statistical power for rare-label and within-family analyses. A post hoc inspection found that the original reference annotations for attentional orientation coding (F4-T3) were less reliable than those for other tasks. These ambiguities reduced the measured accuracy of both VLM and human annotators, complicating interpretation of these results. Future versions should first re-adjudicate these labels with multiple independent annotators. More broadly, developing a larger and more reliable benchmark will require participation from the research community. We encourage HCI researchers to contribute video data licensed for research use, allowing the benchmark to grow over time. Such a shared resource could represent a broader range of annotation practices, enable systematic comparisons across studies, and support more reliable evaluation of VLM annotation.

Finally, our results reflect a single VLM under one prompting and video-input setup. Because model capabilities and prices vary across systems and versions, these findings may not generalize to other VLMs. Future work should replicate the evaluation across models and test its sensitivity to prompting and video sampling.

\section{Conclusion}
We reviewed all 1,702 CHI 2026 full papers, identified 125 that annotated video for research purposes, and developed a five-dimensional taxonomy spanning \emph{analytic purpose, viewpoint, phenomenon, reasoning requirement, and annotation authority}. Grounded in this taxonomy, we constructed a 15-task benchmark and compared VLM-alone annotation, human-alone annotation, and human verification of VLM-generated annotations. VLM-alone annotation was comparable to human-alone annotation on average ($\mathrm{HNS} = 97.0$), but performance varied substantially across tasks. Human verification achieved the highest average accuracy ($\mathrm{HNS} = 121.5$), reduced annotation time by 48.9\%, and cost less than human-alone annotation across the labor rates examined. Our findings show that VLM suitability depends less on viewpoint or temporal horizon than on whether labels can be operationalized from salient visual evidence and general contextual knowledge. VLMs were less reliable for subtle states, specialized conventions, and failures or other low-salience events, whereas verification was most valuable when model errors were quickly recognizable and correctable. By connecting a taxonomy of current practice with this workflow evaluation, our work characterizes VLM capabilities and the conditions for effective human-VLM collaboration, offering researchers concrete questions for deciding \emph{which tasks to automate, when and how to verify model outputs, and how to validate and report VLM-assisted annotations} as auditable research infrastructure.

\bibliographystyle{ACM-Reference-Format}
\bibliography{sample-base}

\appendix
\newpage
\section{Appendix I: Representative examples from each annotation task}
\label{app:examples}

\begin{figure}[h]
    \centering
    \includegraphics[width=0.95\linewidth]{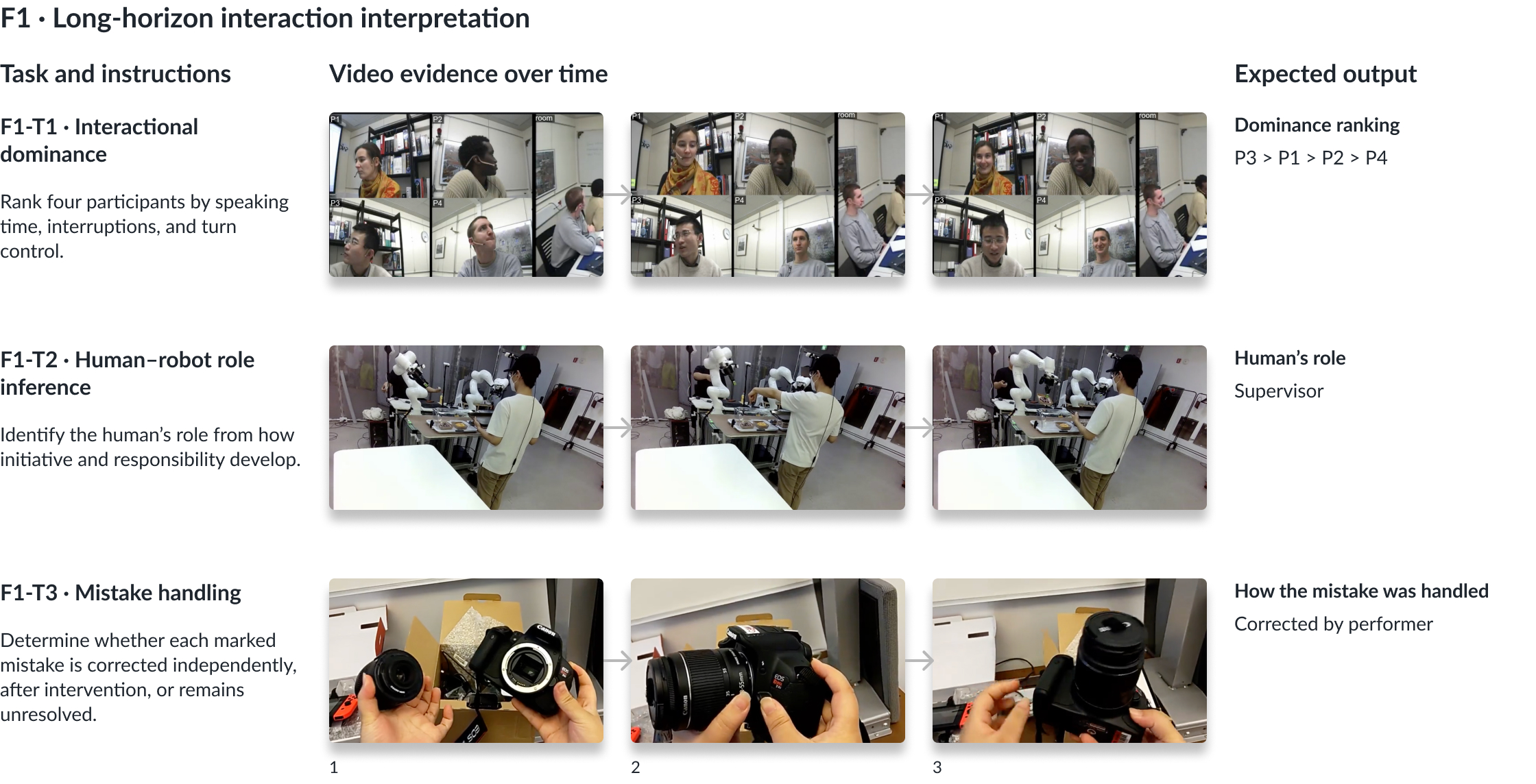}
    \caption{
    Representative examples of the three tasks in the Long Horizon Interaction Interpretation family: interactional dominance, human-robot role inference, and mistake handling. Each row shows the task instructions, representative video evidence over time, and the corresponding expected annotation output.}
    \label{fig:f1_tasks}
    \Description{Representative examples of the three tasks in the Long Horizon Interaction Interpretation family: interactional dominance, human-robot role inference, and mistake handling. Each row shows the task instructions, representative video evidence over time, and the corresponding expected annotation output.}
\end{figure}

\begin{figure}
    \centering
    \includegraphics[width=0.95\linewidth]{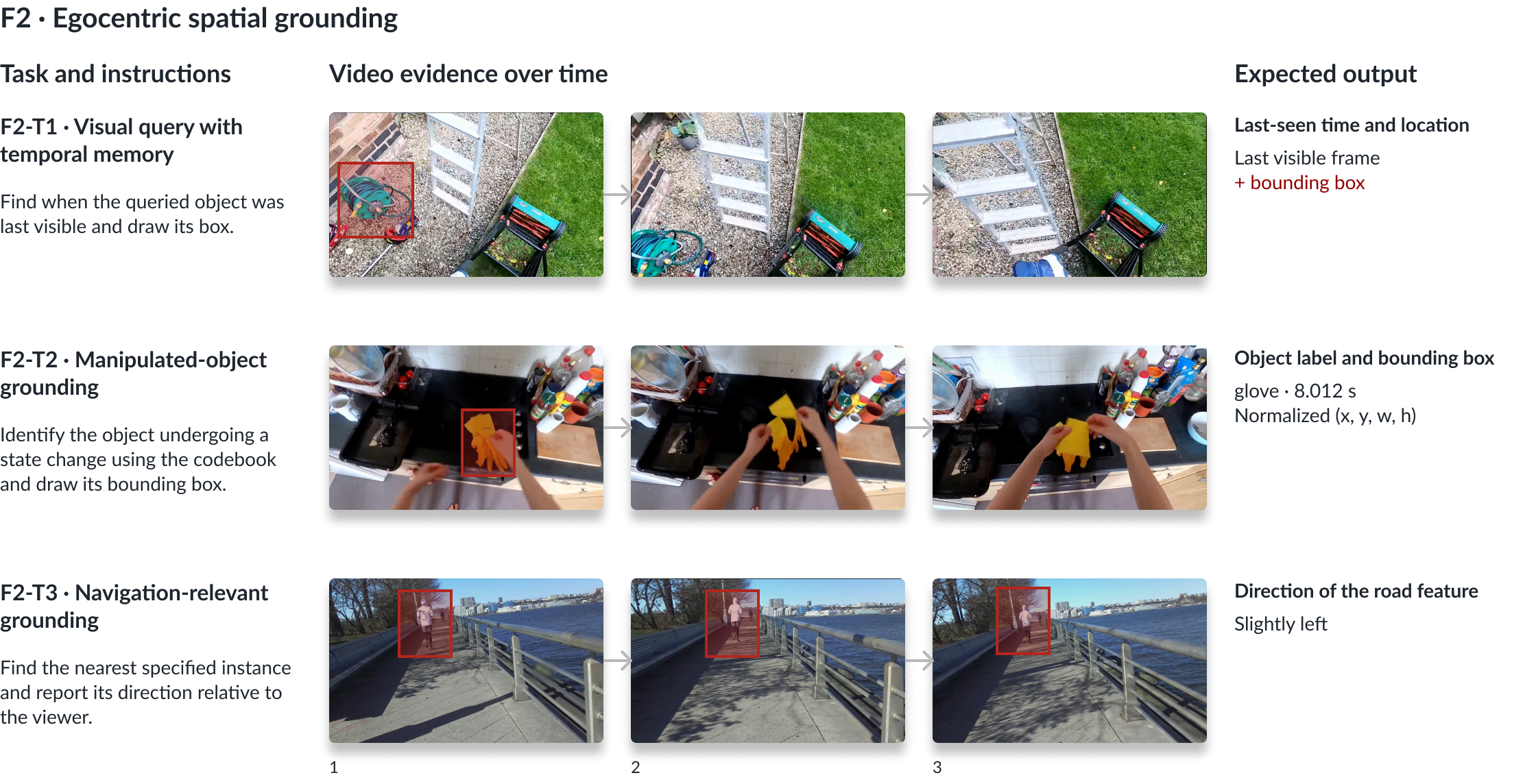}
    \caption{
    Representative examples of the three tasks in the Egocentric Spatial Grounding family: visual query with temporal memory, manipulated object grounding, and navigation-relevant grounding. Each row shows the task instructions, representative video evidence over time, and the corresponding expected annotation output.
    }
    \label{fig:f2_tasks}
    \Description{Representative examples of the three tasks in the Egocentric Spatial Grounding family: visual query with temporal memory, manipulated object grounding, and navigation-relevant grounding. Each row shows the task instructions, representative video evidence over time, and the corresponding expected annotation output.}
\end{figure}

\clearpage
\begin{figure}
    \centering
    \includegraphics[width=0.95\linewidth]{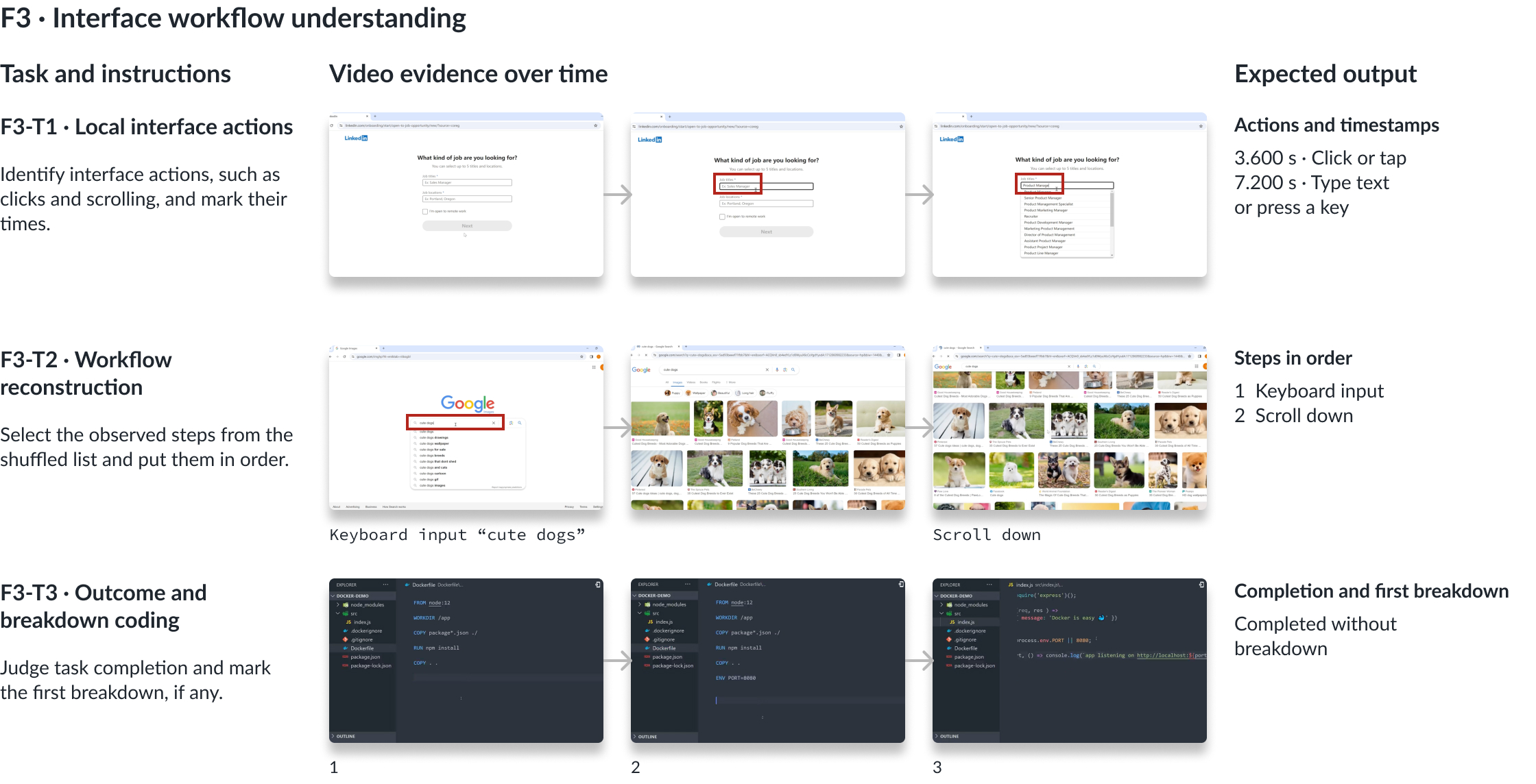}
    \caption{
    Representative examples of the three tasks in the Interface Workflow Understanding family: local interface actions, workflow reconstruction, and outcome and breakdown coding. Each row shows the task instructions, representative video evidence over time, and the corresponding expected annotation output.
    }
    \label{fig:f3_tasks}
    \Description{Representative examples of the three tasks in the Interface Workflow Understanding family: local interface actions, workflow reconstruction, and outcome and breakdown coding. Each row shows the task instructions, representative video evidence over time, and the corresponding expected annotation output.}
\end{figure}

\begin{figure}
    \centering
    \includegraphics[width=0.95\linewidth]{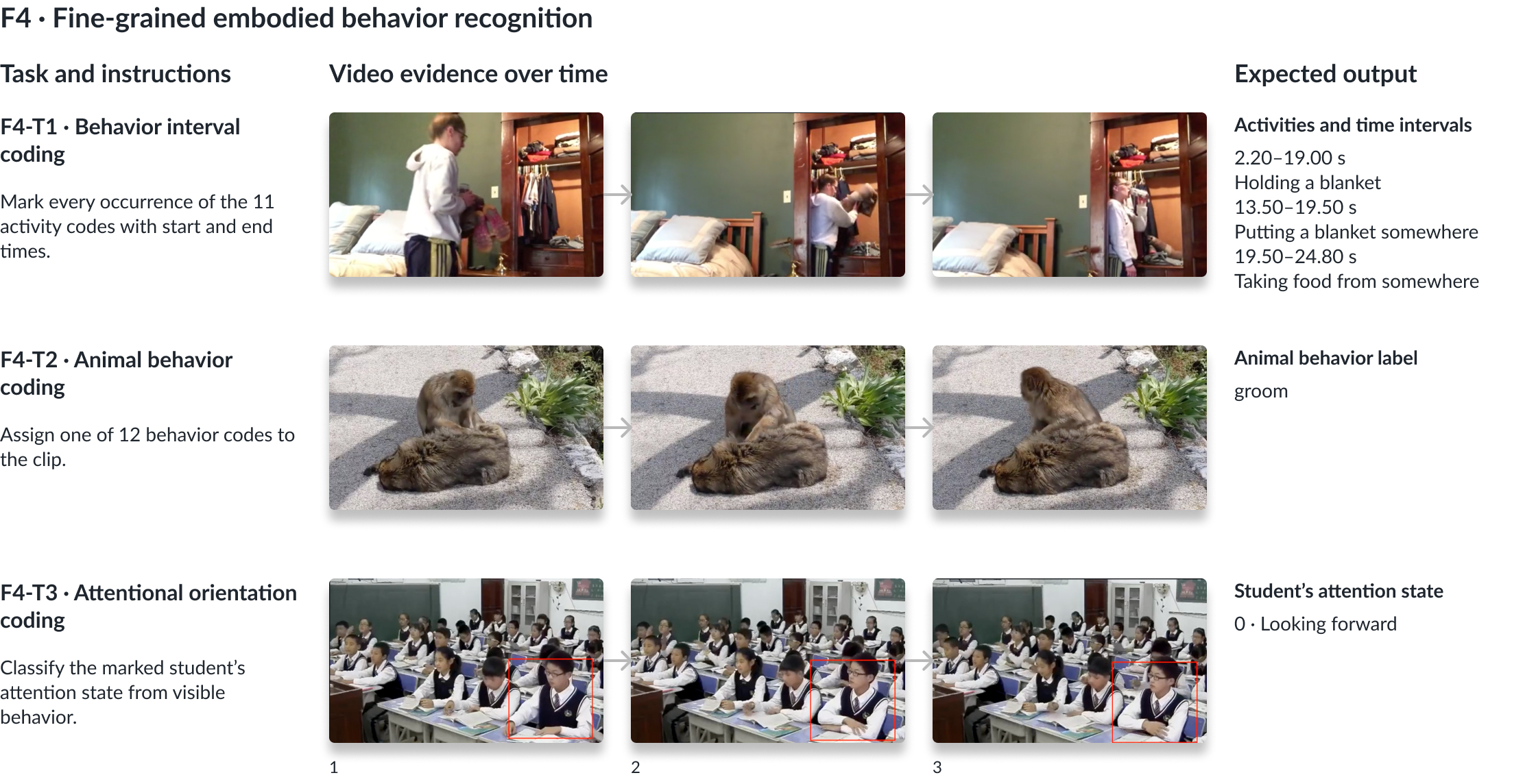}
    \caption{
    Representative examples of the three tasks in the Fine Grained Embodied Behavior Recognition family: behavior interval coding, animal behavior coding, and attentional orientation coding. Each row shows the task instructions, representative video evidence over time, and the corresponding expected annotation output.
    }
    \label{fig:f4_tasks}
    \Description{Representative examples of the three tasks in the Fine Grained Embodied Behavior Recognition family: behavior interval coding, animal behavior coding, and attentional orientation coding. Each row shows the task instructions, representative video evidence over time, and the corresponding expected annotation output.}
\end{figure}

\clearpage
\begin{figure}
    \centering
    \includegraphics[width=\linewidth]{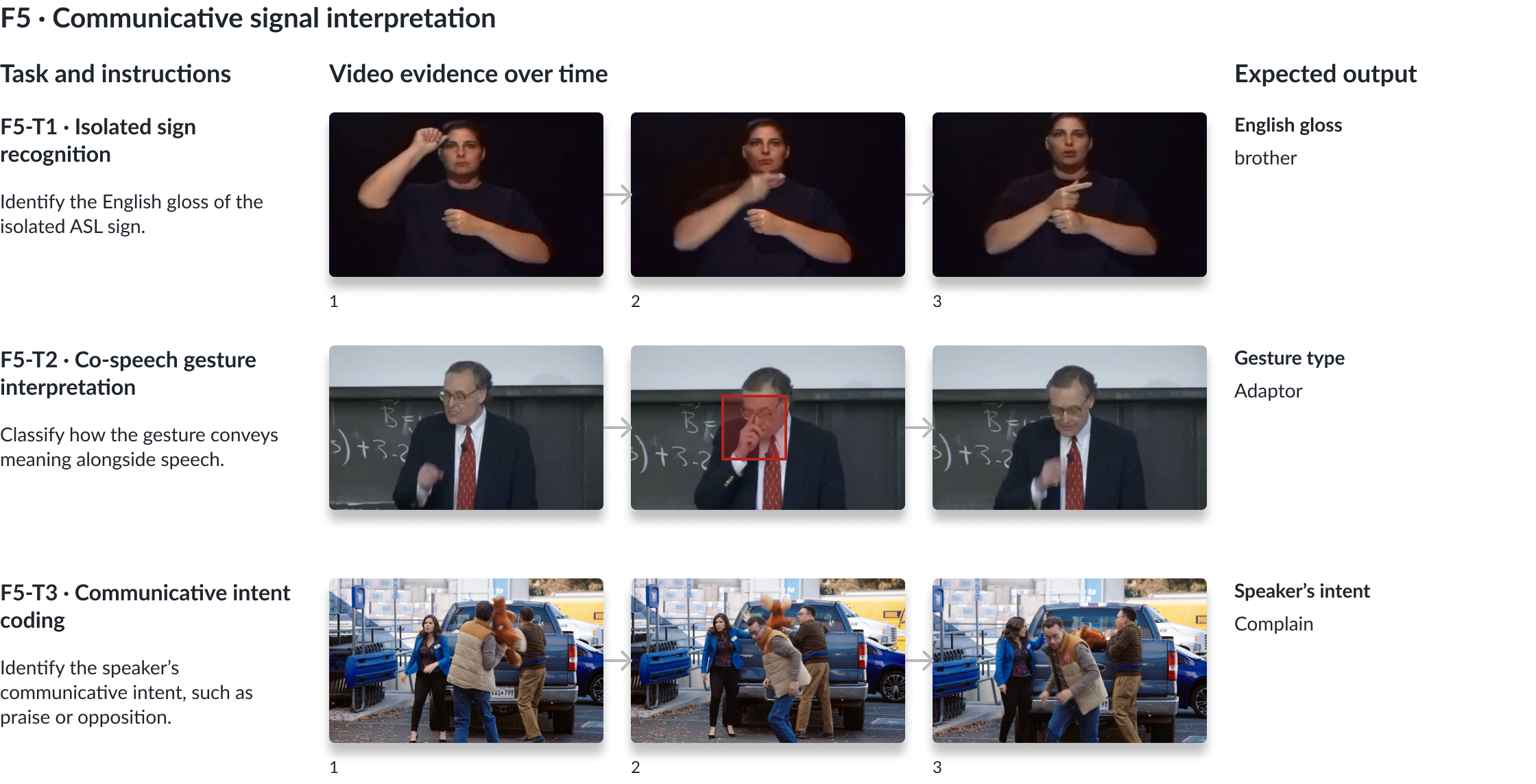}
    \caption{
    Representative examples of the three tasks in the Communicative Signal Interpretation family: isolated sign recognition, co-speech gesture interpretation, and communicative intent coding. Each row shows the task instructions, representative video evidence over time, and the corresponding expected annotation output.
    }
    \label{fig:f5_tasks}
    \Description{Representative examples of the three tasks in the Communicative Signal Interpretation family: isolated sign recognition, co-speech gesture interpretation, and communicative intent coding. Each row shows the task instructions, representative video evidence over time, and the corresponding expected annotation output.}
\end{figure}

% ==================================================================
% Appendix A — Extended Inter-Annotator Agreement
% ==================================================================
\section{Appendix II: Inter-Annotator Annotation Consistency}
\label{app:agreement}

Table~\ref{tab:agreement} reports pairwise annotation consistency on each task's primary metric, computed on the items labeled by all three annotators (the two human annotators H$_1$ and H$_2$, and the VLM). The first three columns are computed in the unassisted annotation mode. The final column reports human--human agreement in the verification mode, where both annotators reviewed the same VLM proposals.

\begin{table*}[h]
\centering
\caption{Pairwise agreement on each task's primary metric between the two human annotators ($H_1$ and $H_2$) and the VLM.}
\Description{Pairwise annotation consistency for the 15 tasks among two human annotators and the VLM, together with human-human consistency when both humans verified the same VLM proposals. Values use each task's primary metric. Verification increases human-human consistency on 11 tasks, including workflow reconstruction from 0.761 to 0.925 and navigation-relevant grounding from 0.571 to 0.905.}
\label{tab:agreement}
\resizebox{0.8\textwidth}{!}{%
\begin{tabular}{@{}lrrrr@{}}
\toprule
Task & $H_1{\leftrightarrow}H_2$ & $H_1{\leftrightarrow}\mathrm{VLM}$ & $H_2{\leftrightarrow}\mathrm{VLM}$ & $H_1{\leftrightarrow}H_2$ (verify) \\
\midrule
Interactional dominance (F1-T1) & 0.767 & 0.767 & 0.800 & 0.867 \\
Human--robot interaction inference (F1-T2) & 0.700 & 0.750 & 0.550 & 0.667 \\
Mistake handling (F1-T3) & 0.762 & 0.714 & 0.667 & 0.714 \\
Visual query with temporal memory (F2-T1) & 0.619 & 0.714 & 0.714 & 0.900 \\
Manipulated-object grounding (F2-T2) & 0.850 & 0.750 & 0.850 & 0.905 \\
Navigation-relevant grounding (F2-T3) & 0.571 & 0.286 & 0.476 & 0.905 \\
Local interface actions (F3-T1) & 0.676 & 0.749 & 0.729 & 0.827 \\
Workflow reconstruction (F3-T2) & 0.761 & 0.750 & 0.764 & 0.925 \\
Outcome and breakdown coding (F3-T3) & 0.680 & 0.380 & 0.280 & 0.760 \\
Behavior interval coding (F4-T1) & 0.696 & 0.739 & 0.775 & 0.887 \\
Animal behavior coding (F4-T2) & 1.000 & 0.850 & 0.850 & 0.905 \\
Attentional orientation coding (F4-T3) & 0.667 & 0.667 & 0.524 & 0.714 \\
Isolated sign recognition (F5-T1) & 1.000 & 0.350 & 0.350 & 1.000 \\
Co-speech gesture interpretation (F5-T2) & 0.714 & 0.667 & 0.476 & 0.840 \\
Communicative intent coding (F5-T3) & 0.650 & 0.800 & 0.550 & 0.900 \\
\bottomrule
\end{tabular}}
\vspace{2pt}
\end{table*}

\newpage
% ==================================================================
% Appendix B — Dataset-Level Monetary Cost Breakdown
% ==================================================================
\section{Appendix III: Task-Level Monetary Cost Breakdown}
\label{app:cost}

\begin{figure*}[h]
\centering
\includegraphics[width=\textwidth]{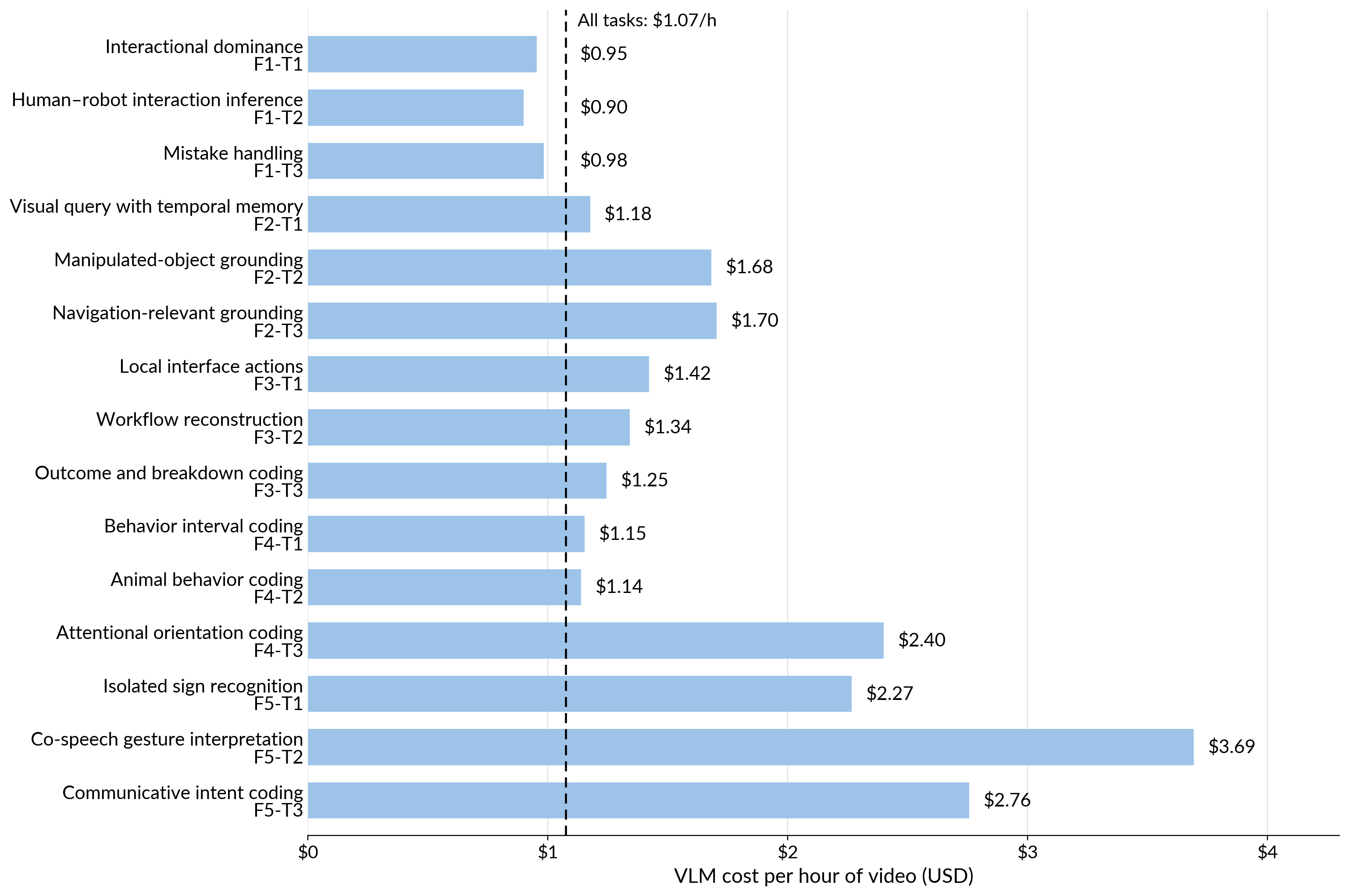}
\caption{Task-level VLM annotation cost breakdown across the 15 benchmark tasks. Per-hour cost varies with clip length because the fixed prompt overhead is amortized over fewer video tokens in short-clip tasks.}
\Description{Horizontal bar chart showing VLM annotation cost per hour of video for the 15 tasks, with a dashed vertical line marking the overall average of 1.07 dollars per hour. Task-level costs range from 0.90 dollars for human-robot interaction inference to 3.69 dollars for co-speech gesture interpretation. Short-clip tasks generally cost more per hour because fixed prompt content is distributed across fewer video tokens.}
\label{fig:vlm-cost}
\end{figure*}

Figure~\ref{fig:vlm-cost} reports the complete dataset-level VLM cost breakdown referenced in Section~\ref{sec:cost}.

\end{document}